%% file: 00_main.tex
\documentclass{article}

\usepackage{iclr2027_conference,times}

\usepackage[T1]{fontenc}
\usepackage[scaled=0.85]{DejaVuSansMono}

\usepackage[utf8]{inputenc} 
\usepackage[T1]{fontenc}    
\usepackage[breaklinks]{hyperref}       
\usepackage{url}            
\usepackage{booktabs}       
\usepackage{amsfonts}       
\usepackage{nicefrac}       
\usepackage{microtype}      
\usepackage{tikz}
\usepackage[table]{xcolor}
\usepackage{lipsum}
\usepackage{tabularx}
\usepackage{array}
\usepackage{url}
\usepackage{graphicx}
\usepackage{wrapfig}
\usepackage{subcaption}
\usepackage{booktabs} 
\usepackage{multirow}
\usepackage{diagbox}
\usepackage{adjustbox}
\usepackage{enumitem}
\usepackage{listings}
\usepackage{pifont}

\usepackage{minted}
\usepackage{tcolorbox}
\tcbuselibrary{breakable}

\input{math_commands}

\definecolor{kellygreen}{rgb}{0.3, 0.73, 0.09}
\definecolor{alizarin}{rgb}{0.82, 0.1, 0.26}

\title{
LLMs are General Asynchronous Agents
}

\author{%
  George Yakushev$^{\star\,\dagger\,\lozenge}$\,
  Denis Mazur$^{\star\,\ddag}$\,
  Vladimir Bartenev$^{\dagger\,\triangle}$\,
  Vyacheslav Zhdanovskiy$^{\dagger}$\,
  \AND
  Timofey Byzov$^{\dagger\,\triangle}$\,
  Vladimir Kaurkin$^{\lozenge}$\,
  Vadim Pastushenko$^{\triangle\lozenge}$
  \\
}

\begingroup
\makeatletter
\renewcommand\@makefnmark{}
\footnotetext{\hspace{-15px}$^\star$Equal Contribution,
$^\dagger$Yandex,
$^\ddag\,$Together AI,
$^\lozenge$HSE university,\\
$^\triangle$Yandex School of Data Analysis,
Correspondence to: \texttt{yakushev-ga@yandex-team.ru}\,.}
\makeatother
\endgroup

\iclrfinalcopy 
\begin{document}

\maketitle

\begin{abstract}
Modern LLMs are increasingly capable as autonomous agents, but they follow sequential interaction cycles: read, think, reply or call tools, repeat.
Many real-world use cases are not sequential: voice assistants, embodied agents, and monitoring systems receive new inputs while they think or perform another task.
Modern LLMs address this with specialized architectures for voice interaction and video streams, VLAs for robot control, asynchronous tool calling for API usage, and others.
In this work, we generalize from different asynchronous tasks to general asynchronous agents that can adapt to different types of concurrency.
To achieve this, we develop an asynchronous LLM framework that lets users (or the agents themselves) define inference coroutines with overlapping memory states.
We showcase that Qwen 3.x models are capable of asynchronous operation for streaming video understanding, videogames, and monitoring, without task-specific training.

\end{abstract}

\input{01_intro}

\input{02_background}

\input{03_method}

\input{04_experiments}

\input{05_discussion}

\input{06_statements}

\bibliography{main}
\bibliographystyle{iclr2027_conference}

\input{07_appendix}

\end{document}

%% file: math_commands.tex
\usepackage{amsmath,amsfonts,bm}

\def\eqref#1{equation~\ref{#1}}

\def\1{\bm{1}}

\DeclareMathAlphabet{\mathsfit}{\encodingdefault}{\sfdefault}{m}{sl}
\SetMathAlphabet{\mathsfit}{bold}{\encodingdefault}{\sfdefault}{bx}{n}



%% file: 01_intro.tex
\vspace{-12px}
\section{Introduction}\label{sect:introduction}
\vspace{-7px}

Large language models (LLMs) are becoming increasingly capable autonomous agents, enabled by recent advances in reinforcement learning, tool use, and inference-time compute~\citep{kimik2openagentic,challenging_bigbench_solved_with_cot_Suzgun2022ChallengingBT,beeching2024scalingtesttimecompute}. 
Modern LLMs can solve problems that require hours of uninterrupted reasoning, programming, and tool use~\citep{jimenez2024swe,Schick2023ToolformerLM, pmlr-v202-gao23f}\nocite{Shen2023HuggingGPTSA,Qin2023ToolLLMFL,azerbayev2024llemma, wang2024mathcoder, li2024chainofcode,openai_arc_prize_o3,kwa2025measuring_metr}.
To solve these complex tasks, LLM agents follow a Thought-Action-Observation loop~\citep{Yao2022ReActSR}: instead of solving the problem in one go, the agent reasons and defines an action such as running code, then observes the outcome (e.g., error traceback) to inform its next step.

However, not all use cases allow for turn-based problem solving: a real-time voice assistant needs to listen while thinking and handling interruptions~\citep{audio_moshi}, a self-driving car must quickly adjust to changes in traffic~\citep{zhou2024hazard}, and even a fully virtual monitoring agent needs to react quickly when the operating system has issues~\citep{chen2025aiopslab,tang2026devops_gym}.
Human ``agents'' do this naturally, sometimes without thinking, because we evolved to process continuous information streams and react to new stimuli~\citep{eriksen1979information,wessel2017globality}.

However, artificial LLM agents are not naturally asynchronous as they were built for sequences and trained on turn-based interaction.
The current state of the art treats each application with concurrency as a separate research problem.
Real-time voice and video assistants are explicitly built for concurrent thinking and listening~\citep{audio_moshi,wang2024a_fullduplex,lin2025asyncvoiceagentrealtimeexplanation,huang2026wan} and adjust when interrupted~\citep{audio_gpt4o,cao2025interruption}.
For embodied agents, Vision-Language-Action (VLA) models~\citep{vla_Driess2023PaLME} often follow a dual actor-thinker architecture~\citep{song2025humeintroducingsystem2thinking,tan2025think} so they can react to stimuli while thinking.
The latest programming harnesses let the user ``steer'' the agent with extra inputs during reasoning~\citep{openai_mid_turn_steering}; others use asynchronous function calling or even multiple parallel sub-agents~\citep{ning2024skeletonofthought,ginart2410asynchronous,zheng2025parallelr1}.
Currently, each of these research areas designs and trains agents for concurrency in their own task-specific manner.

\begin{figure}
    \centering
    \vspace{-30px}
    \includegraphics[width=0.999\linewidth]{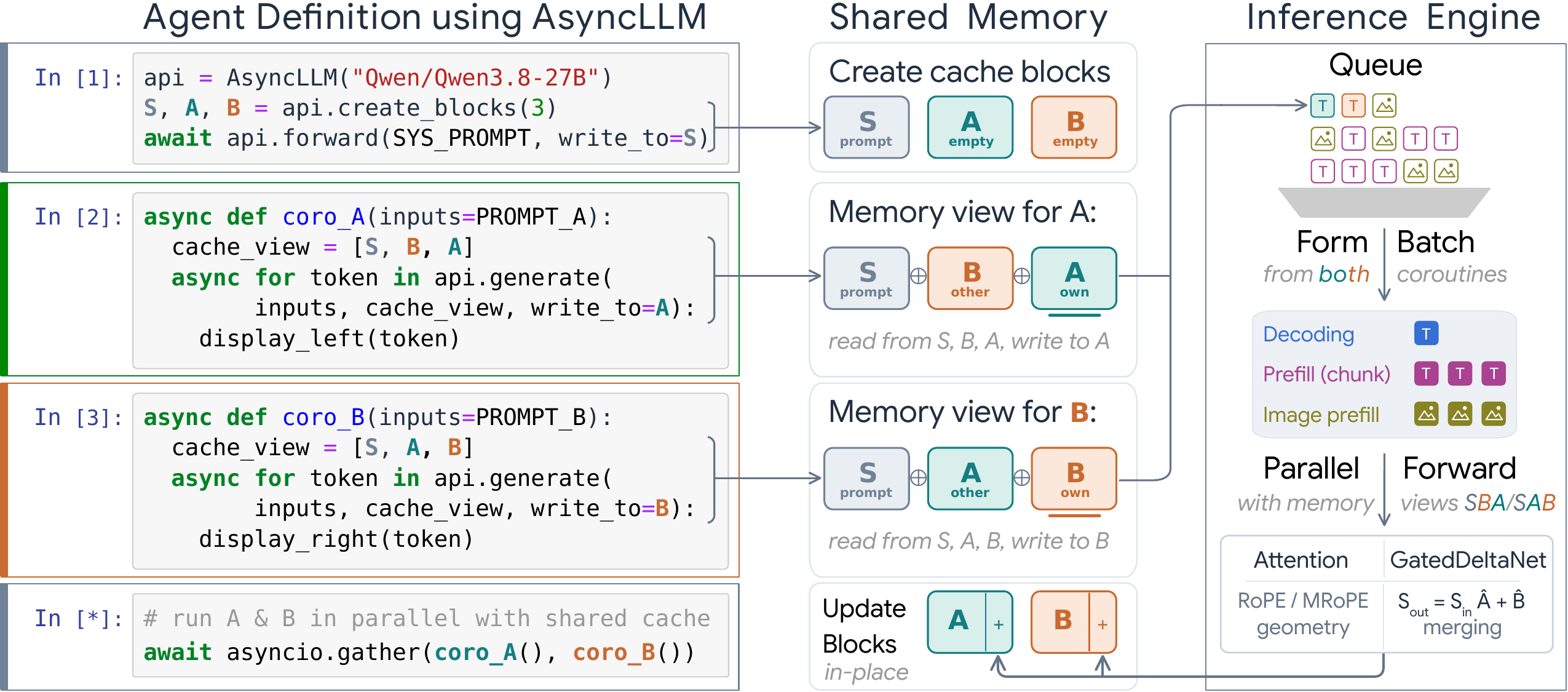}
    \vspace{-15px}
    \caption{An example of AsyncLLM agent design with two parallel sub-agents solving a math task with shared memory: \textbf{(left)} the agent is defined as a set of coroutines that write into shared cache blocks; \textbf{(middle)} the cache blocks are arranged in ``views'' that define how coroutines see each other's work; \textbf{(right)} the engine groups coroutines into batches for efficient inference. Details in Section~\ref{sect:method}.}
    \label{fig:teaser}
    \vspace{-10px}
\end{figure}

In this work, we study whether LLM agents can be made \textit{generally asynchronous}, similarly to how they are general tool users and few-shot learners.
We hypothesize that, because modern LLMs learn to mimic human reasoning at some stage of their training, they may be able to imitate human asynchrony with proper framing.
To test this, we design a general framework for defining asynchronous LLM agents without task-specific fine-tuning.
Since communicating via text would be slow for many real-time applications, we leverage direct memory sharing~\citep{hogwild_inference,zheng2025parallelr1} and extend its algorithms to support hybrid and multimodal language models.
We adopt the popular async/await programming model via \texttt{asyncio}~\citep{vanrossum2012asyncio}\nocite{syme2011f,claessen1999poor,baker1977incremental} to support concurrent LLM inference with asynchronous inputs and outputs, as depicted in Figure~\ref{fig:teaser}.

AsyncLLM lets the developer (or the agent itself) define multiple asynchronous coroutines that execute concurrently. 
Each coroutine writes to its own ``memory block'' and can access other coroutines'  outputs using attention ``views''.
This allows the developer to define coroutines that see each other's progress in real time and handle new environment I/O without waiting for current reasoning to finish.
Our inference engine automatically groups concurrent inference requests for efficient batched GPU inference with shared memory, where every coroutine has its own view on the same cache blocks.
Our framework lets practitioners adapt existing state-of-the-art models to new concurrency scenarios or combine multiple scenarios that would otherwise require specialized data and expensive fine-tuning.
The three main contributions of this work can be summarized as follows:

\begin{itemize}[leftmargin=1em]
    \vspace{-8px}\item We propose \textrm{AsyncLLM}, a general framework for \textit{training-free} asynchronous LLM agents in the async/await programming model. 
    Our framework extends asynchronous programming primitives to define parallel LLM inference coroutines with overlapping memory states.
    \vspace{-3px}\item We describe an algorithm for parallel GPU inference with shared memory states (attention KVs, GDN recurrent states), allowing multiple instances of the same LLM to run concurrently while seeing each other's progress in real time. Our algorithm supports hybrid and multimodal LLMs.\footnote{\url{https://github.com/dvmazur/async_llm}}\
    \vspace{-3px}\item We test the generality of AsyncLLM by constructing asynchronous agents for streaming video understanding, videogame environments, and system monitoring, based on the same family of Qwen3.x VLMs without task-specific fine-tuning. 
    Our experiments demonstrate that modern LLMs can use this programming model to define and modify their own coroutines in the same inference loop. 
    While this capability is not yet reliable, our results suggest that future LLM generations may create self-adapting agents from environment descriptions.
\end{itemize}\vspace{-3px}

%% file: 02_background.tex
\vspace{-10px}
\section{Background}\label{sect:background}
\vspace{-8px}
Recent works have come up with asynchronous agents across vastly different research areas. In this section, we overview several of these areas and draw parallels in how they handle concurrency.

\textbf{Voice assistants}~\citep{audio_palm,audio_speechgpt,audio_gpt4o,gemini_live,claude_voice_mode_2025} communicate with users in real time using either an
ASR\nocite{first_asr_davis1952,kaldi,wav2vec,whisper}-LLM-TTS~\nocite{first_tts_umeda1968,ssps2009,wavenet,tacotron,tacotron2,waveglow,hifigan,tortoisetts} pipeline or, more recently, a multimodal foundation model~\citep{audio_qwen,audio_moshi,mini_omni,llama_omni}\nocite{salmonomni1,salmonomni2,omniflatten}.
However, spoken conversation requires more than multimodality~\citep{human_conversation_corpus_study,proactive_voice_2020,MAHMOOD2025103406}: natural speakers ask questions while thinking, interrupt each other, and read nonverbal cues as they talk.
This becomes even more pronounced in group conversation or talking while working together in a shared coding environment~\citep{Flamino2025_testing_the_limits,agent_brainstorming,daryanto2026humanhumanaitriadicprogramminguncovering,welter2025developerpairsaicopilots}.
To maintain natural conversations, modern voice assistants work in full-duplex mode, i.e., listen, think, and speak concurrently~\citep{audio_moshi,wang2024a_fullduplex,veluri-etal-2024-beyond}\nocite{wu2026chronologicalthinkingfullduplexspoken} with a slower background ``thinker''~\citep{lin2025asyncvoiceagentrealtimeexplanation,zhang2026liberating,zou2026ltsvoiceagentlistenthinkspeakframeworkefficient,wu2026silentthoughtmodelinginternal,huang2026duplexomnirealtimelisteningseeing}.
Advanced voice assistants have modules that detect interruptions (``barge-in'') to pause and adjust the response~\citep{selfridge-etal-2013-continuously,zhao-etal-2015-incremental,cao2025interruption}.

\textbf{In streaming video understanding}~\citep{mun2019streamlineddensevideocaptioning}\nocite{wang-etal-2025-videollm,VideoLLM-online}, the model must keep up with real-time video to detect industrial incidents~\citep{anomalygpt,yang2025assistpdaonlinevideosurveillance,yuan2024towards}, assist driving~\citep{huang2025online,zheng2026driveagent} or comment on sporting events~\citep{mkhallati2023soccernet,yang2026magevl}.
As video signals are denser than audio, models typically cannot process every frame in real-time.
To combat this, recent works train lightweight ``probes'' that determine which frames can be skipped~\citep{wang-etal-2025-videollm,kim2025egospeaklearningspeakegocentric,ding2025streammind,yang2026magevl}, and keep a small window of recent video frames and compress past events using text descriptions~\citep{streamingVLM}, hidden representations~\citep{qian2024streaminglongvideounderstanding}, or retrieval~\citep{ning2025livevlm}.
Streaming video models can watch and reason concurrently to reduce response delays~\citep{guan2026video,qian2025dispiderenablingvideollms}.
The mechanisms used in these models are similar to the ones used in full-duplex voice assistants with interruption handling, but the probe is used not for voice interruptions but to detect changes in traffic situation~\citep{probe_sdc,zheng2026driveagent}, handle GUI pop-ups, or react to user's nonverbal cues~\citep{smartkom, patapati2025geneca}.
Full video assistants~\citep{liu2024streamchat,wang2026streambridge,huang2026wan} and GUI computer use agents~\citep{lin2025showui,li2026agentcomputerobservationinterfacesenable} use similar techniques to process multiple input streams simultaneously.

\textbf{Embodied agents}~\citep{vla_Ahn2022SayCan}\nocite{vla_Driess2023PaLME,vla_Mon-Williams2025ELLMER,vla_Wang2023Voyager,vla_Jiang2023VIMA} use Vision-Language-Action models~\citep{vla_Brohan2023RT2, vla_openVLA, vla_survey} that process visual and text inputs and choose actions for a physical system they control. Most VLAs focus on a certain type of robotic system, such as mobile manipulators~\citep{intelligence2025pi_} or humanoid robots~\citep{bjorck2025gr00t} and require fine-tuning to adapt to a new type~\citep{wang2025vlaadaptereffectiveparadigmtinyscale,sun2026vla}, while several more recent VLAs are trained to support several different embodiments~\citep{team2025gemini_robotics,luo2026being}.
Similar to voice and video assistants, embodied agents operate in an inherently asynchronous world and need to quickly adapt to interruptions and changes in the environment~\citep{zhou2024hazard,gonzalez2025robotouille,borate2025llm,cao2025interruption}.
Similar to full-duplex assistants, embodied agents think and act concurrently and use dedicated subroutines for processing unexpected interruptions~\citep{song2025humeintroducingsystem2thinking,liu2025vita}\nocite{fang2025dualvla,tan2025think,wei2026libra}.

\textbf{Virtual \& Computer Use agents}\nocite{hu2025agents,sager2026comprehensive} can control terminal shells~\citep{llm_manage_linux,llm_autonomous_attack,merrill2026terminal}, web browsers~\citep{hilton2021webgpt,deng2023mind2web,zheng2024gpt_web,zhou2024webarena,koh2024visualwebarena}, virtual environments~\citep{llm_minecraft,ma2024large_starcraft_ii,almeida2026jev}, desktop~\citep{llm_oscopilot,xie2024osworld,hong2024cogagent} or mobile operating systems~\citep{zhang2023appagentmultimodalagentssmartphone,you2024ferret}. Their design varies between applications: a terminal agent has text-only inputs, a videogame agent requires vision and audio, and browser agents have both GUI~\citep{koh2024visualwebarena,hong2024cogagent} and text-based inputs~\citep{hilton2021webgpt,zhou2024webarena}. Similarly, the need for asynchrony varies from one application to another, but follows the same general patterns. A system monitoring agent~\citep{qi2023loggpt,shetty2024building,tang2026devops_gym}\nocite{jha2025itbench} needs to process a continuous stream of logs from a running system to detect problems such as memory leaks or runaway processes, which is similar to streaming video understanding. Modern coding assistants allow users to alter an already running request via mid-turn steering~\citep{openai_mid_turn_steering} and ask by-the-way questions~\citep{anthropic_claude_code_commands_btw}. Though the exact steering mechanism is not disclosed, it follows the same pattern of concurrency as voice assistant interruptions.

\textbf{Parallel reasoning and tool use.} Parallel to application-specific asynchronous agents, several recent lines of research use concurrency in parallel LLM reasoning~\citep{efficient_reasoning_survey,ning2024skeletonofthought,Wang2022SelfConsistencyIC}, asynchronous function calling~\citep{ginart2410asynchronous,gim2024asynchronousllmfunctioncalling,kim2024llmcompiler}, and recursive sub-agents~\citep{zhu-etal-2024-redel,zhang2025rlm}. In parallel reasoning, multiple LLM instances reason on the same problem together, solving subtasks or debating ideas~\citep{ning2024skeletonofthought,jin2025learningpromisescalinglanguage}\nocite{yang2026multiverse,wen2025parathinker,du2024improving}. Others apply a similar technique to overlap thinking with reading a long input~\citep{tong2025streamingthinkerlargelanguagemodels}, writing a response~\citep{yakushev2025asynchronous}, or waiting for tool calls~\citep{ginart2410asynchronous,gim2024asynchronousllmfunctioncalling}. Recent works found that giving parallel sub-instances real-time access to each other's thoughts allows them to coordinate faster~\citep{hogwild_inference,zheng2025parallelr1} and ensure safety while reasoning~\citep{yakushev2025asynchronous,wang2026beyond}.

\vspace{-1px}The applications we reviewed differ in modalities and deployment requirements, but they use similar patterns of concurrency. Parallel inference streams are used in both full-duplex visual assistant and for subtasks in parallel reasoning. Both streaming video understanding and system monitoring benefit from event probes. Both assistants and system monitors launch subroutines to handle interruptions. These agents use custom inference software that implements task-specific parallelism and communication and need to adapt the model for their setup. In this work, we propose a framework that generalizes between these applications without the need for task-specific training and inference.

%% file: 03_method.tex


\newpage
\section{Asynchronous Agents}\label{sect:method}
\vspace{-7px}
We design AsyncLLM around the async/await programming model using Python \texttt{asyncio} standard library~\citep{vanrossum2012asyncio}, where concurrency is defined through coroutines and synchronization primitives.
These coroutines run concurrent LLM inference while communicating through composable memory states (CacheBlocks) that contain attention KV caches and Gated Delta Network recurrent states~\citep{yang2025gated} for a slice of tokens.
Unlike prior works, AsyncLLM groups coroutines into batched GPU execution automatically, allowing users to focus on application logic.

\input{resources/fig_method_code_example}

Figure~\ref{fig:method_code_example} shows how these components work together for an asynchronous agent that reasons about its task and simultaneously provides the user with the running summary of its progress, similar to interleaved or asynchronous reasoning~\citep{xie2025interleaved,yakushev2025asynchronous}.
The agent uses three cache blocks: one for the prompt, one for private reasoning, and one for the user-facing response.
This way, the ``thinker'' coroutine can write new thoughts into the reasoning block while the ``writer'' summarizes them in the response block.
The implementation consists of two coroutines: a ``thinker'' that produces the reasoning trace and a ``writer'' that summarizes it.
Since the writer needs to see the current reasoning progress, its cache view (L17) contains the thinker block before its own summary, reusing the memory state within.
In turn, the thinker does not need to see the writer's output to reason about the problem, so its cache view only includes the problem and its own reasoning.
The thinker never explicitly communicates tokens to the writer. Instead it only notifies it about a finished paragraph, so the writer can summarize the progress by accessing the shared thinking block.

We organize the rest of this section as follows: Section~\ref{sect:method_framework} defines the AsyncLLM framework in more detail, Section~\ref{sect:method_math_generalization} describes the algorithms for quickly reconstructing memory states from consecutive cache blocks by extending prior work on attention cache manipulation~\citep{hogwild_inference}; Section~\ref{sect:method_engine_details} covers efficient GPU inference with attention views.


\vspace{-5px}
\subsection{AsyncLLM Programming Model}\label{sect:method_framework}
\vspace{-5px}

Every forward pass in AsyncLLM writes its memory state to a \texttt{CacheBlock} that contains
the model's internal memory state for a continuous chunk of tokens. For modern hybrid transformers, this corresponds to KV caches for attention layers and state transitions for Gated Delta Nets (GDN).
Every forward pass writes to a single cache block, but can ``see'' multiple other cache blocks at once in arbitrary order.
We refer to these compositions as cache views, as depicted in Figures~\ref{fig:teaser}~\&~\ref{fig:method_code_example}.
Attending to multiple cache blocks is mathematically equivalent to attending to a single conventional KV cache containing the same tokens if those blocks were encoded sequentially. If multiple coroutines update their blocks concurrently while looking at each other, the resulting memory state is not equivalent to any sequential inference, but it remains legible to the LLM as we show below. This memory-based parallelism lets asynchronous agents run parallel sub-routines that synchronize instantly.
In contrast, if coroutines communicated with tokens, they would have to re-encode previously generated tokens every time the memory view changes, e.g. whenever the ``thinker'' in Figure~\ref{fig:method_code_example} generates a new token.


The agent updates its cache blocks by running LLM forward passes and saving the internal state to the designated cache block.
In the example above, \texttt{api.forward} (prefill) and \texttt{api.generate} use the same underlying forward pass algorithm that attends to a cache view and updates the provided cache block (\texttt{write\_to}).
AsyncLLM engine runs concurrent inference forward passes from different coroutines in the same batch, as we discuss in Section~\ref{sect:method_engine_details}. Additional cache blocks can be created via \texttt{api.create\_block()}, emptied with \texttt{cache\_block.clear()}, and merged into one via \texttt{api.merge\_blocks(A, B)}, which is equivalent to attending to both blocks but with less overhead.


Asynchronous LLM applications require the agent to react to signals: voice assistant interruptions, streaming video events, browser GUI pop-ups and, others.
In our framework this can be expressed by combining LLM inference with \texttt{asyncio} synchronization primitives: events, locks, queues, and so on.
For instance, consider a streaming video understanding agent that selects important frames with a ``probe'', then notifies a background thinking coroutine about the event. In \texttt{asyncio / async\_llm}, this can be expressed with an \texttt{asyncio.Event} that notifies the background thinker of an update. When multiple event-describing coroutines compile their descriptions into a shared report, they can use an \texttt{asyncio.Lock} to ensure the output is not garbled when merging.

\vspace{-8px}
\subsection{Inference with Multiple Cache Blocks}\label{sect:method_math_generalization}
\vspace{-5px}

Next, we describe how AsyncLLM runs forward passes while attending to multiple cache blocks (\texttt{cache\_view}). Recall that every cache block contains KV vectors and GDN recurrent states for a contiguous slice of tokens across all model layers. For traditional attention KV layers, we could combine KV caches by rotating every subsequent cache block's keys to their new positions according to Rotary Position Embeddings (RoPE,~\citealp{su2021roformer}). However, that would require rearranging all past KVs for every forward pass, which would slow down inference. \citet{hogwild_inference} show that, for full attention with RoPE, one can compute attention without rotating previous KV blocks. Instead, they rotate only the current attention queries, keeping previous keys and values as-is.
Intuitively, consider the attention dot product $\langle \rho(q, i), \rho(k, j)\rangle$, where $\rho(\cdot, \cdot)$ applies RoPE rotation that encodes the query position $i$ or the key position $j$. It can be rewritten as follows:
\begin{equation}\langle \rho(q, i), \rho(k, j)\rangle = \langle \rho(q, i - j), \rho(k, 0)\rangle, \: \text{where} \: \rho(k, 0) = k\end{equation}
The resulting algorithm keeps past KV caches on fixed 0-based positions (0, 1, 2, \dots) and computes attention by rotating the current queries relative to each block, producing equivalent attention outputs.

However, modern LLMs are not limited to traditional RoPE attention layers: most state-of-the-art open-weight models are hybrids where full attention layers are interleaved with either sliding window attention~\citep{gemmateam2026gemma4technicalreport,abadji2026lagunam1xs2technicalreport}\nocite{beltagy2020longformer} or, more frequently, linear attention or Delta Network variants~\citep{qwen35blog,kimiteam2026kimik3openfrontier,zeng2026glm}\nocite{schlag2021linear,yang2025gated}.
Additionally, multimodal LLMs use multimodal rotary embeddings (MRoPE)~\citep{wang2024qwen2_mrope} for images, audio and video inputs. In AsyncLLM, we adopt the attention manipulation from~\citet{hogwild_inference} for full attention layers and propose new algorithms for linear and multimodal attention.

\textbf{Concurrent Linear Attention and Gated Delta Nets.} Unlike full attention, linear attentions such as GDN and KDA\nocite{zhang2025kda} use a fixed-size recurrent state $S_t \in \mathbb R^{d_v \times d_k}$ for each head. On every new token, the model predicts learned projections, e.g. $q_t, k_t, v_t, \alpha_t, \beta_t$ for GDNs, then updates $S_t$ and outputs $o_t$:
\begin{equation}
S_t = S_{t-1}\left(\alpha_t\left(I - \beta_t k_t k_t^{\top}\right)\right) + \beta_t v_t k_t^{\top}, \quad\quad\quad\quad\quad o_t = S_t q_t
\end{equation}

When computing $S_t$ with multiple memory blocks in \texttt{cache\_view}, we reformulate the problem from per-token to per-block computation. Each CacheBlock contains a transition from the initial state before the block $S_{T_0}$ to the state after the block $S_{T_{max}}$. For modern Delta Network variants, this transition is affine in the incoming state: linear in $S_{T_0}$ up to an additive, input-dependent term. For convenience, let us rewrite the GDN update using auxiliary matrices $A_t, B_t$:
\begin{equation}
S_t=S_{t-1}A_t+B_t,\qquad
A_t=\alpha_t\!\left(I-\beta_t k_tk_t^\top\right),\qquad
B_t=\beta_t v_tk_t^\top .
\label{eq:gdn_reformulation}
\end{equation}
Similarly, we can rewrite the update for $S_t$ over several consecutive steps in terms of $A_t, B_t$:
\begin{equation}
S_{t}=S_{t-1} A_{t} + B_{t} = \left(S_{t-2}A_{t-1}+B_{t-1}\right)A_{t} + B_{t} = S_{t-2}\, A_{t-1} A_{t} + B_{t-1}A_{t} + B_{t}
\end{equation}
\begin{equation}
S_{t}=S_{t-2}\, \hat A_{t,2} + \hat B_{t,2}, \;\;\text{where}\;\;
\hat A_{t,2}=A_{t-1}A_t,\quad
\hat B_{t,2}=B_{t-1}A_t+B_t .
\end{equation}
Or more generally, we can pack $n$ consecutive state updates as $S_t = S_{t-n}\, \hat A_{t, n} + \hat B_{t, n}$, with
\begin{equation}
\hat{A}_{t,n}
=
\prod_{i=t-n+1}^{t} A_i,
\qquad
\hat{B}_{t,n}
=
\sum_{i=t-n+1}^{t}
B_i
\prod_{j=i+1}^{t} A_j , \;\text{where empty products equal}\;I
\label{eq:block_generalized}
\end{equation}

For AsyncLLM inference with multiple cache blocks, we represent each GDN head within the block with a pair of matrices $(\hat A,\hat B)$ from the above equation that summarize all steps within this block as per Eq. (\ref{eq:block_generalized}). Then, the GDN state after two consecutive cache blocks $L$ \& $R$ can be composed as:
\begin{equation}
S_{LR} = (S_0 \hat A_L + \hat B_L) \hat A_R + \hat B_R, \quad\quad (\hat A_L,\hat B_L)\circ(\hat A_R,\hat B_R)
=(\hat A_L\hat A_R,\;\hat B_L\hat A_R+\hat B_R).
\end{equation}
This requires $O(N_\text{blocks})$ instead of $O(N_\text{tokens})$  matrix operations per head and can be done just-in-time for a given forward pass. As coroutines progress and new tokens are added, the matrices $\hat A, \hat B$ in each block are updated incrementally per Eq. (\ref{eq:block_generalized}) with a linear transform for $\hat A$ and an affine transform for $\hat B$.
If another memory view orders these blocks differently, the same matrices are multiplied in a different order.
This lets AsyncLLM inference engine compose attention views with hybrid LLMs using GatedDeltaNets.
If the blocks are arranged in a different order, as in Figure~\ref{fig:teaser}, the final memory state is computed by multiplying the same pairs of matrices in a different order.
Other linear attention layers follow the same computation with slightly different definitions of the original $A_t, B_t$. For instance,  KDA replaces the scalar decay $\alpha_t$  in Eq. (\ref{eq:gdn_reformulation}) with a diagonal gate $\mathrm{Diag}(\alpha_t)$, and the rest holds verbatim. In other words, other popular linear attention variants also support memory views.

\textbf{Concurrent multimodal attention.} When processing image and video data, modern MLLMs use multimodal or multi-dimensional positional embeddings to encode a given image patch's row and column indices along with its position in the video or image set.
MRoPE partitions key (or query) dimensions into temporal, height, and width sections and rotates each section based on a given token's frame index (temporal), grid row (height) and grid column (width) indices.
If AsyncLLM attends to a cache block that contains image tokens, we rotate the current query to match the relative position in the temporal dimension and keep the two spatial axes unchanged. However, there is one more caveat that affects how cache blocks are stacked together. Since an image has multiple visual tokens for every temporal frame, a cache block with $T$ mixed visual and text tokens spans fewer than $T$ temporal positions, and it affects how the next block should be rotated. To account for this, we explicitly maintain per-block position spans instead of rotating by token count.


Combined with prior work on concurrent attention, this allows us run to inference on modern hybrid MLLMs with arbitrary cache views. Linear attention layers compose block-wise affine transitions, while full-attention layers use query rotation with MRoPE correction. The remaining MLP layers are invariant to cache views and can be computed normally. We discuss additional architecture variants in Appendix~\ref{app:architectures_compatibility}. The remaining challenge is batching multiple coroutines on the same device.

\vspace{-8px}
\subsection{Scheduling and Inference Engine}\label{sect:method_engine_details}
\vspace{-5px}
We build the AsyncLLM inference engine over mini-SGLang\footnote{Based on \url{https://github.com/sgl-project/mini-sglang}}, a minimal implementation of the SGLang framework~\citep{zheng2024efficiently}. Our reference implementation consists of three main components: \textbf{i)} the front-end that implements the \texttt{asyncio}-based AsyncLLM interface from Section~\ref{sect:method_framework}, \textbf{ii)} a balanced scheduling algorithm that lets reaction coroutines complete quickly without being clogged, and \textbf{iii)} efficient batched inference kernels with memory views based on Section~\ref{sect:method_math_generalization}.

\textbf{Scheduling for asynchronous agents.} AsyncLLM gathers incoming forward pass requests from all coroutines into a shared queue and forms batches for parallel execution. Both prefill and generation requests are mapped to the same batched LLM forward algorithm where requests of different lentghs and cache views can run in the same batch. To avoid situations where a quick response coroutine gets ``clogged'' by larger background requests, our engine forms balanced batches from all active coroutines. We employ chunked prefilling~\citep{agrawal2023sarathi_chunked_prefill}, i.e. mapping long prefills into multiple mathematically equivalent forward passes that can be batched with decoding for interactivity.

\textbf{Batched Inference.} We take advantage of modern LLM frameworks using Paged Attention~\citep{kwon2023efficient,zheng2024efficiently}, i.e. splitting the KV cache into ``pages''.
This lets us implement CacheBlocks as a light-weight data structure that holds references to attention pages and GDN affine transitions and implements update, merge, and clear operations.
When running a forward pass on a given batch, we use the existing mini-SGLang kernels for everything except inner Attention and GatedDeltaNet (after projections).
We track which cache views are needed for the active batch and feed them into the corresponding kernels: full attention with query rotation and MRoPE modifications, and GDN kernels that use the algorithm in Section~\ref{sect:method_math_generalization} to construct the GDN state, then run Flash Linear Attention~\citep{yang2024fla}\nocite{zhang2025kda,chen2026attnres,yang2025path,yang2024deltanet,zhang2024gsa}. We discuss additional implementation details in Appendix~\ref{app:implementation_details} and evaluate inference latency and throughput under different synthetic workloads in Appendix~\ref{app:gpu_throughput_experiments}.

%% file: resources/fig_method_code_example.tex
\begin{figure}[b!]
    \centering
    \vspace{-10px}
    \begin{minted}[fontsize=\small, linenos, numberblanklines=false]{python}
import asyncio, async_llm
api = async_llm.AsyncLLM(MODEL_NAME, **config)
prompt_block, thinker_block, writer_block = await api.create_blocks(3)
paragraph_finished = asyncio.Event()              # thinker-writer synchronization
await api.forward(PROMPT, write_to=prompt_block)  # prefill prompt into prompt_block

async def thinker_coro():
    async for token in api.generate(
        "<think>\n", cache_view=[prompt_block, thinker_block], write_to=thinker_block,
    ):  # this coroutine updates thinker_block that the writer will summarize --^
        if token == "\n\n":
            paragraph_finished.set()  # notify the writer

think_in_background = asyncio.create_task(thinker_coro())
while not think_in_background.done():
    await paragraph_finished.wait()  # wait for background thoughts
    paragraph_finished.clear()
    async for token in api.generate(
        "...</think> Summary:", cache_view=[prompt_block, thinker_block, writer_block],
        write_to=writer_block, stop_at="\n",  # writer sees --^ thoughts in real-time
    ):
        send_to_user(token)     # e.g. speak via TTS
    \end{minted}
    \vspace{-10px}
    \caption{Example AsyncLLM agent for implementing parallel reasoning and speaking (simplified).}
    \label{fig:method_code_example}
    \vspace{-15px}
\end{figure}







%% file: 04_experiments.tex
\newpage
\vspace{-10px}
\section{Experiments}\label{sect:experiments}
\vspace{-7px}

The core idea of this work is that LLM agents can be made \textit{generally} asynchronous without training, and that memory view manipulations from Section~\ref{sect:method_math_generalization} let them react to asynchronous inputs. To better isolate each individual claim, we begin by evaluating individual text and image changes in Section~\ref{sect:experiments_sanity_check_individual_changes}. Next, we evaluate AsyncLLM agents on real-time streaming video in Section~\ref{sect:experiments_streaming_video_understanding}, interactive agents for videogames in Section~\ref{sect:experiments_videogames}, and text-based system monitoring in Section~\ref{sect:experiments_monitoring}.

\begin{figure*}[t!]
    \centering
    \vspace{-30px}
    \includegraphics[width=0.49\linewidth]{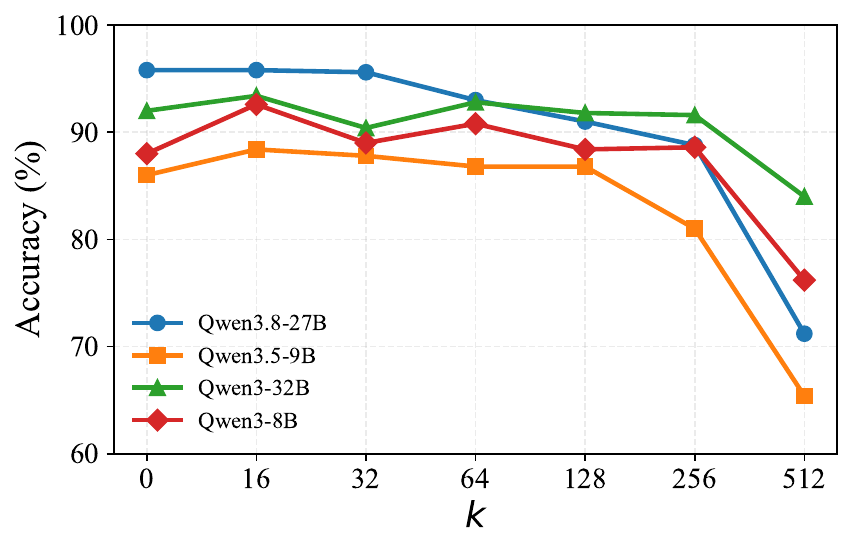} \includegraphics[width=0.49\linewidth]{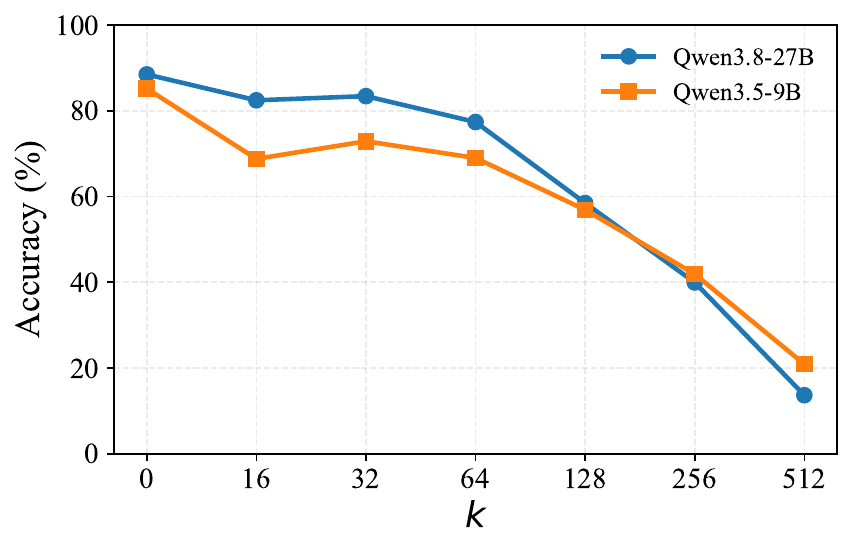}
    \vspace{-12px}
    \caption{
    Accuracy with one asynchronous input; the x-axis denotes when the input arrived (step~\#).
    \textbf{(left)} text clarifications on MATH-500-Sharded, \textbf{(right)} image changes on ShardedVQA (Section~\ref{sect:experiments_sanity_check_individual_changes}). 
    }
    \label{fig:experiments_sanity_checks_extra_inputs}
    \vspace{-15px}
\end{figure*}

\vspace{-8px}
\subsection{Sanity Checks: Single Asynchronous Input}\label{sect:experiments_sanity_check_individual_changes}
\vspace{-5px}

We first test the simplest asynchronous setting: mid-reasoning user clarifications. The agent begins solving a math or image-QA task, then receives a missing detail or correction while thinking. We evaluate an AsyncLLM agent equivalent to AsyncReasoning~\citep{yakushev2025asynchronous}, extended to hybrid LLMs using Section~\ref{sect:method_math_generalization}, on hybrid-attention Qwen 3.5+ models~\citep{qwen35blog}\nocite{qwen38blog}.

\textbf{Asynchronous text inputs.} We evaluate hybrid LLMs on MATH-500-Sharded~\citep{yakushev2025asynchronous}, where each of 500 math problems~\citep{hendrycksmath2021} is split into an incomplete prompt and a clarification. The clarification is provided after $k$ reasoning steps. We report these evaluations in Figure~\ref{fig:experiments_sanity_checks_extra_inputs} (left) and provide non-hybrid AsyncReasoning for reference.

\textbf{Asynchronous image changes.} Next, we evaluate whether VLM-based agents can revise ongoing reasoning when visual evidence changes. We construct 513 image pairs from existing visual QA and math datasets.
For each problem, we edit the input image to add realistic ``errors'' and use
the original image as the ``revised'' version. The errors are constructed so that the problem cannot be solved correctly from the erroneous image. Dataset construction is described in Appendix~\ref{app:sharded_visual_reasoning_dataset}. The agent starts working with the erroneous (edited) image. After $k$ decoding steps, we replace it with the corrected image and add a notification that the input has changed.

Figure~\ref{fig:experiments_sanity_checks_extra_inputs} summarizes our results: AsyncLLM agents based on Qwen 3.x hybrid MLLMs can react to asynchronous inputs in both visual and text modalities. The visual agent shows trends similar to text interruptions, but the accuracy drops somewhat faster: upon closer inspection, we found that this is explained by the fact that visual tasks, on average, require less reasoning than MATH-500, and larger $k$ sometimes arrive when the agent has already produced the answer. Still, our results demonstrate that AsyncLLM agents can react to inputs while reasoning. In subsequent sections, we use this ability to process visual and text streams in more complex applications.

\vspace{-8px}
\subsection{Streaming Video Understanding}\label{sect:experiments_streaming_video_understanding}
\vspace{-7px}

We switch from a single image change to monitoring a continuous video stream and evaluate training-free Streaming Video Understanding. We evaluate on the SoccerNet-Caption~\citep{mkhallati2023soccernet} sports commentary dataset using the streaming evaluation protocol from~\citet{ding2025streammind}, and on the ProactiveVideoQA mixed video benchmark using the official protocol~\citep{wang2025proactivevideoqa}.

The AsyncLLM agent consists of five concurrent components: \textbf{1) the event probe} runs every frame and determines if it contains a new event worth describing.
If the probe triggers, the image pair (cache block) is passed to \textbf{2) the background thinking thread} that reconstructs video events from frames.
At the same time, \textbf{3) the output probe} checks if the reasoning contains a new event. If it does, the \textbf{4) description writer} generates the event description from the thinker's internal state, and a final non-LLM \textbf{5) output compiler} coroutine gathers the results. Our agent uses \textit{training-free probes}: instead of training a separate sub-module as in~\citet{ding2025streammind,yang2026magevl}, we run the base LLM as-is with a pre-filled prompt that determines important frames in a single forward pass\nocite{almeida2026jev}. We provide a detailed agent description and prompts in Appendix~\ref{app:streaming_video_understanding_agent}.

\begin{figure*}[t!]
    \centering
    \vspace{-20px}
    \captionof{table}{
    Streaming video understanding evaluation with AsyncLLM Qwen 3.x models and a non-AsyncLLM Mage-VL streaming model: \textbf{(left)} SoccerNet results in streaming video understanding protocol, \textbf{(right)} ProactiveVideoQA evaluation across four subsets and totals, using the metrics from the original evaluation protocol (all higher is better). See details and discussion in Section~\ref{sect:experiments_streaming_video_understanding}.
    }
    \label{tab:experiments_streaming_video_understanding}
    \vspace{-5px}

\renewcommand{\arraystretch}{1.25}
\setlength{\tabcolsep}{3pt}
\begin{tabular}{lccc|ccccccc}
\toprule
\multirow{2}{*}{Agent \& Model} & \multicolumn{3}{c|}{\small SoccerNet (streaming)} & \multicolumn{5}{c}{\small ProactiveVideoQA PAUC ($\omega{=}0.5$)} & \!\!{\small{Trigger$_\text{Acc}$}}\!\! & {\small TimVal} \\

\cmidrule(lr){2-4} \cmidrule(lr){5-9} \cmidrule{10-10} \cmidrule{11-11}

& {\small AUROC} & {\small{Trigger$_\text{Acc}$}}\! & {\small TimVal} & {\small WEB} & {\small EGO} & {\small TV} & {\small VAD} & {\small ALL} & {\small ALL} & {\small ALL} \\
\midrule
Qwen3.5-9B   & 0.677 & 62.82 & 39.62 & 0.493 & 0.563 & 0.638 & 0.358 & 0.541 & 52.99 & 17.19 \\
Qwen3.8-27B  & 0.608 & 54.88 & 33.27 & 0.501 & 0.482 & 0.616 & 0.366 & 0.504 & 52.45 & 16.82 \\
Q3.6-35B-A3B & 0.651 & 62.55 & 37.46 & 0.476 & 0.578 & 0.624 & 0.344 & 0.539 & 52.30 & 16.37 \\
\midrule
Mage-VL      & 0.555 & 52.79 & 27.87 & 0.323 & 0.538 & 0.390 & 0.284 & 0.428 & 43.27 & 10.03 \\
\bottomrule
\end{tabular}
\vspace{-15px}
\end{figure*}

We evaluate AsyncLLM with three Qwen 3.x models and compare it against Mage-VL~\citep{yang2026magevl}, a recent model trained specifically for video streaming (see Appendix~\ref{app:streaming_video_understanding_extra_evals} for hyperparameter tuning). We follow the standard evaluation protocol for both benchmarks: event probe ROC AUC, TriggerAcc (fraction of correctly
predicted events), and TimVal (which balances speak and silence correctness). On ProactiveVideoQA, we also report Proactive AUC with the recommended response delay weight (PAUC $\omega{=}0.5$) and report other $\omega$ in Appendix~\ref{app:streaming_video_understanding_extra_evals}. Table~\ref{tab:experiments_streaming_video_understanding} summarizes our results: AsyncLLM with Qwen3.5+ models outperforms the more specialist Mage-VL on both benchmarks. While the official evaluation protocol for SoccerNet does not measure description quality, the larger Qwen3.5+ models also provide more accurate descriptions on ProactiveVideoQA. We see this not a weakness of Mage-VL that was trained with different priorities, but a positive side-effect of using generalist MLLMs in AsyncLLM. Finally, we report inference speed in Table~\ref{tab:experiments_throughput}.

\vspace{-5px}
\subsection{Interactive Agents for Video Games}\label{sect:experiments_videogames}
\vspace{-5px}

\begin{figure*}[b!]
    \centering
    \vspace{-17px}
    \includegraphics[height=120px,width=0.485\linewidth]{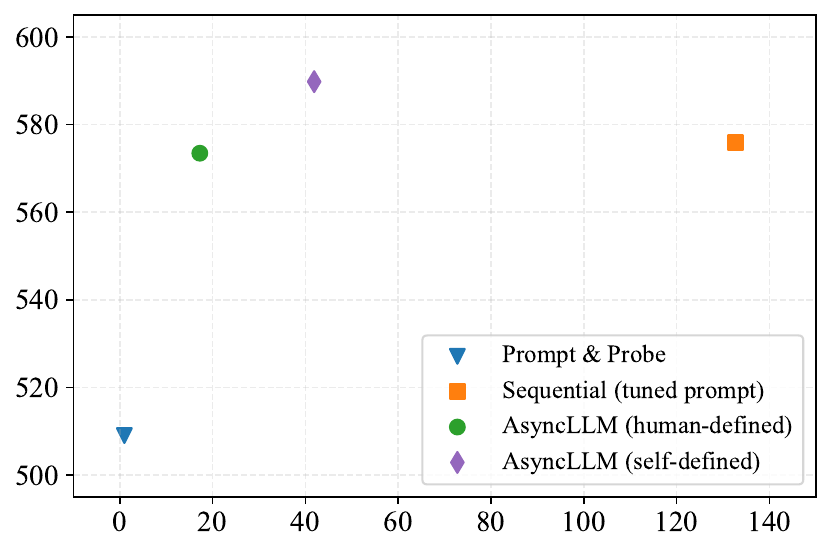} \includegraphics[height=120px,width=0.485\linewidth]{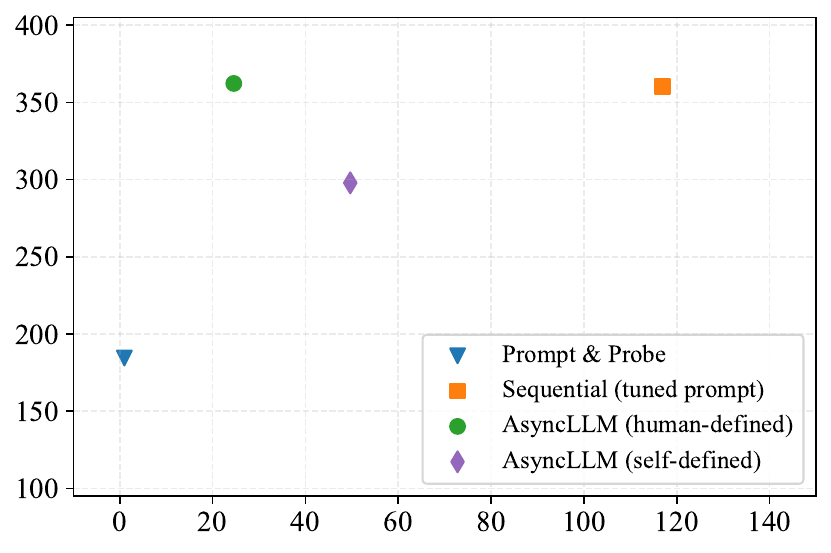}
    \vspace{-8px}
    \captionof{figure}{
    AsyncLLM Qwen3.6-35B-A3B evaluation on VizDoom environments \textbf{(left)} DoomHealthGathering and \textbf{(right)} DoomDeadlyCorridor; the y-axis is the average reward and the x-axis is the mean delay (forward passes) from observation to taking an action, averaged over 100 episodes.
    }
    \label{fig:experiments_videogames_doom}
    \vspace{-15px}
\end{figure*}

Next, we move from passive monitoring to agents that affect their environment. We evaluate on two ViZDoom scenarios based on the videogame Doom~\citep{Kempka2016ViZDoom,wydmuch2018vizdoom}\nocite{towers2026gymnasium}: HealthGathering and DeadlyCorridor (frame skip 4).
We modify the AsyncLLM agent architecture from the previous section: instead of describing the video stream, the background thinker module determines the course of action (e.g. ``turn right until you are aiming at the rightmost enemy'') and a faster action subroutine converts it into frame-by-frame actions with an early-exit probe that determines when the action can be picked.
We compare against two baseline agents: sequential and probe-only.
Before interacting, the agents are prompted with the environment rules and action space and allowed to reason about strategy.
Then, on each frame, they observe two latest frames from the environment and are probed to choose action.
We prompt the sequential agent to write short single-paragraph reasoning to avoid overthinking.
We also evaluate a faster probe-only agent that runs one forward pass per frame with a pre-written template\footnote{``Based on \dots, the next action is: \_\_\_'', after which we take a valid action token with the highest probability.}.
The results in Figure~\ref{fig:experiments_videogames_doom} show that the AsyncLLM agent can react much faster than a sequential agent based on the same model while preserving the gains from reasoning.
This demonstrates that AsyncLLM can generalize existing VLMs to interactive videogame agents without training. That said, our results only demonstrate the basic capability, not state-of-the-art performance for these specific environments.
We also experiment with letting \textbf{agents define their own coroutines} based on the environment and AsyncLLM API (see Appendix~\ref{app:experiments_self_definition}), which shows good initial results on HealthGathering but inferior on DeadlyCorridor.

\newpage
\vspace{-5px}
\subsection{System Monitoring Agents}\label{sect:experiments_monitoring}
\vspace{-5px}

For our next evaluation, we analyze the AsyncLLM agent for system monitoring~\citep{qi2023loggpt,chen2025aiopslab,tang2026devops_gym}. In this setup, the agent works with an input stream of system logs from a running application and detects anomalies, such as memory leaks or runaway file descriptors. We use the Monitoring subset from the DevOps-Gym benchmark~\citep{tang2026devops_gym} consisting of 34 tasks.
The original setup only measured accuracy with several types of anomalies such as network latency spikes or system handle leaks.
We modify our evaluation setup to account for how quickly the system responds: we let the agents perceive logs in real-time and measure, on average, how soon the agent answered and how many inference steps it used. To make our results GPU-agnostic, we report the number of forward passes. This is equal to the number of output tokens for the sequential baseline but counts two tokens generated in the same batch as one forward pass (see in Table~\ref{tab:experiments_throughput} for throughput reference). The AsyncLLM agent has, on average, 2.06 active coroutines for this task.

The AsyncLLM monitoring agent follows the same architecture as in Streaming Video Understanding except for a different input format: instead of images, the agent receives chunks of logs split by lines up to 1 second or 1024 characters long (at least one line). We compare AsyncLLM against three alternatives that use the same paradigm: a baseline sequential agent with the same model that gives the answer after the whole sequence, a sequential agent with an early-answer tool after every log chunk (same prompt as AsyncLLM), and a faster baseline agent that checks whether it should reason on a given chunk of logs or skip it entirely.
We use an additional prompt paragraph that ensures the agent only worries about resource usage issues and limits its reasoning to keep up with the current events. Without that, we found that Qwen 3.6-35B-A3B was overthinking unrelated issues and fell behind real-time logs resulting in ${<}10\%$ accuracy. We apply the same anti-distraction prompt to AsyncLLM and baselines. Table~\ref{tab:experiments_monitoring} summarizes our results: for both models, the asynchronous monitoring agents can detect problems with comparable accuracy, but significantly faster than comparable sequential agents, using same concurrency principles as the streaming video agent in Section~\ref{sect:experiments_streaming_video_understanding}.

\begin{figure}[t]
    \centering
    \vspace{-25px}
    \begin{minipage}[t]{0.525\linewidth}
        \captionof{table}{DevOps-Gym Monitoring results with anti-distraction prompt and streaming eval for Section~\ref{sect:experiments_monitoring}.}
        \vspace{-5px}
        \label{tab:experiments_monitoring}
        \centering
        \renewcommand{\arraystretch}{1.095}
        \setlength{\tabcolsep}{4pt}
        \begin{tabular}{lccc}
            \toprule
            \multirow{2}{*}{Agent} & Accuracy & \# forward & Mean \% \\
                  &(\%),  $\uparrow$  & passes $\downarrow$ & events $\downarrow$ \\
            \midrule
            \rowcolor{gray!20}Q3.6-35B-A3B & 61.76 & 8452 & 100\\
            + early answer  &47.06 & 6641 & 58.71\\
            + skip rows     &26.47 & 3686 & 65.86\\
            AsyncLLM        &55.88 & 3837 & 55.49\\
            \midrule
            \rowcolor{gray!20}Qwen3.5-9B & 47.06 & 8726 & 100 \\
            + early answer  & 44.12 & 5749 & 61.21 \\
            + skip rows     & 8.82 & 2204 & 78.23 \\
            AsyncLLM        & 41.18 & 2582 & 44.98 \\
            \bottomrule
        \end{tabular}
    \end{minipage}
    \hfill
    \begin{minipage}[t]{0.465\linewidth}
        \captionof{table}{AsyncLLM inference speed, 1x H200
        (top) decoding throughput on synthetic tasks, (bottom) real-time rate on ProactiveVideoQA.}
        \vspace{-5px}
        \label{tab:experiments_throughput}
        \centering
        \renewcommand{\arraystretch}{1}
        \setlength{\tabcolsep}{7pt}
        \begin{tabular}{lccc}
            \toprule
            Model & 9B & 27B & 35B-A3B \\
            \midrule
            \multicolumn{4}{c}{Throughput$\uparrow$, tokens per second.}\\
            \midrule
            1 coros    & 106 & 43  & 96  \\
            2 coros    & 192 & 80  & 169 \\
            4 coros    & 339 & 144 & 284 \\
            8 coros    & 552 & 239 & 438 \\
            \midrule
            \multicolumn{4}{c}{Average stream seconds$\uparrow$ per GPU-second.}\\
            \midrule
            Video only   & 4.4 & 2.9 & 2.5 \\
            + audio (TV)   & 1.8 & 1.4 & 1.4 \\
            \bottomrule
        \end{tabular}
    \end{minipage}
    \vspace{-15px}
\end{figure}

%% file: 05_discussion.tex
\vspace{-10px}
\section{Discussion and Future Work}\label{sect:discussion}
\vspace{-8px}

In this work, we investigated the ability of large language models to operate as asynchronous agents over a diverse set of environments.
Our results suggest that asynchrony can be treated as a general capability rather than a collection of domain-specific skills.
We found that modern MLLMs already have the capacity to handle concurrent sub-tasks and react to new stimuli without task-specific asynchronous training. 
Similarly to early tool-use agents, they do not always outperform purpose-trained models on their specific domains, but are much easier to adapt to new applications.
To facilitate this adaptation, we formulated AsyncLLM, a framework that lets the user define asynchronous agents in the async/await programming model, and proposed efficient algorithms that extend this capability to modern hybrid and multimodal LLMs.
Our experiments demonstrate that AsyncLLM agents can generalize to asynchronous text and visual updates, streaming video understanding, system monitoring, and even interactive videogame agents. 
This could enable real-time asynchronous APIs that let users run LLMs in asynchronous environments, e.g. by supplying the agent definition through the API. Alternatively, proprietary LLMs can write their own harness from the user's prompt.

This opens two interesting directions for future research: \textbf{i)} training LLMs to be better general asynchronous agents, in the same sense that current LLMs train to be better general tool use agents, and \textbf{ii)} further investigating the agent's ability to improve its own coroutines for the given scenario.

%% file: 06_statements.tex
\section*{Acknowledgements}
Authors would like to thank Max Ryabinin, Alina Shutova and Gleb Rodionov for helpful discussions, brainstorming the experiment scenarios, advice with presentation, and proofreading.

\section*{AI use statement}

In this work, we used generative AI tools for generating, cleaning, and reformatting a partially synthetic dataset, and minor thematic analysis during early prototyping, and we experimented with using generative AI to implement an agent in one of our experiments.
In particular, we used generative AI tools for generating and cleaning synthetic incomplete images for asynchronous image inputs dataset (Appendix~\ref{app:sharded_visual_reasoning_dataset}).
We also used a generative AI API for LLM-as-a-judge evaluation for MATH-500~\citep{hendrycksmath2021} as part of its canonical evaluation protocol.
Appendix~\ref{app:experiments_self_definition} describes the case where we used the generative AI (asynchronous agent) to write its own AsyncLLM definition for the environment it operates in.
We have not used generative AI tools for developing theoretical models or conceptual frameworks, assisting with translation, proposing hypotheses, or providing feedback on research methodology.
Formulating mathematical claims, providing ingredients for these claims, or assisting in writing proofs are not applicable to this work.
Additionally, we used search-augmented generative AI tools to help find additional related works, and we used generative AI coding assistants to draft empty templates for certain tables and plots. We have reviewed all AI-assisted work: we checked the synthetic erroneous images manually and ran VLM inference to ensure that the problem cannot be solved without the corrected image. We also verified the Python plot templates and tables before entering our results. We take responsibility for the final content of this work, including text, claims or artifacts produced with the aid of
generative AI.

\section*{Ethics statement}
This work studies general-purpose asynchronous LLM agents that can observe, reason, and act concurrently. While this can improve responsiveness in applications such as streaming assistants, monitoring systems, and embodied or interactive agents, it also introduces additional risks. Concurrent execution may cause an agent to emit an incorrect or harmful action before slower reasoning or monitoring coroutines can intervene, and asynchronous inputs may change an agent's behavior while other computations are still in progress.
More capable asynchronous agents may also increase the usefulness of LLMs for dual-use applications such as automated system interaction, surveillance, or other continuously operating agents. Additionally, we ran experiments with self-modifying agents that needed sandboxing and strict limits on the tools, files, and external systems that an agent is allowed to modify.
Real-world deployments of AsyncLLM agents in sensitive use cases should include appropriate guardrails, sandboxing, output/action validation, and monitoring, depending on the deployment scenario.

\section*{Reproducibility statement}

To facilitate reproducibility, we provide our reference implementation of AsyncLLM and the sharded VQA dataset in the supplementary materials. We also provide experiment configurations and prompts used for asynchronous agent and dataset creation. The supplementary code and evaluation dataset will be released upon publication.

%% file: 07_appendix.tex
\pagebreak
\appendix

\section{Patterns of Concurrency}
\label{app:concurrency-patterns}

While conducting the experiments for this work, we noticed that most applications, despite their variety in function and modalities, share a set of recurring design patterns.
We distinguish five complementary patterns; a single system may combine several.
These design patterns could be further abstracted away from developers in higher-level frameworks.
Table~\ref{tab:concurrency-patterns} compares representative implementations of these patterns and their training requirements.

\vspace{-8px}
\paragraph{Probes and monitors.}
A probe inspects incoming observations or an ongoing generation and decides
whether to initiate, suspend, or redirect computation. Examples include
speech-initiation prediction, event-gated video processing, and reasoning-safety
monitoring
\citep{kim2025egospeaklearningspeakegocentric,ding2025streammind,wang2026beyond}.
Here, ``probe'' describes a control role: it may be implemented using a trained
classifier, a heuristic, or a prompted frozen model.

\vspace{-8px}
\paragraph{Parallel inference streams.}
Multiple activities progress over overlapping time intervals, potentially at
different rates. Examples include listening while speaking, reasoning while
writing, and slow planning alongside fast action generation
\citep{audio_moshi,yakushev2025asynchronous,song2025humeintroducingsystem2thinking}.
Streams may operate independently or consume one another's intermediate outputs.

\vspace{-8px}
\paragraph{Subroutines and delegation.}
A parent computation launches bounded work---such as a tool call, a reasoning
branch, or a recursive sub-agent---and subsequently incorporates its result
\citep{kim2024llmcompiler,zhu-etal-2024-redel,zhang2025rlm}.
This pattern becomes concurrent when sibling subroutines overlap or the parent
continues before a result arrives. Recursive decomposition alone does not imply
asynchronous execution.

\vspace{-8px}
\paragraph{Interruptions and incremental inputs.}
New observations, user corrections, or tool results arrive after computation
has started and influence its subsequent behavior. The system may pause,
revise, resume, or replace an ongoing response
\citep{cao2025interruption,gim2024asynchronousllmfunctioncalling,liu2024streamchat}.
Implementations differ in whether updates are handled between model calls or
incorporated directly into an active generation.

\vspace{-8px}
\paragraph{Shared and evolving context.}
Concurrent activities communicate through messages, intermediate text, or
shared model state. Concurrent attention exposes partial generations to other
workers, while streaming memory retains and retrieves earlier observations
\citep{hogwild_inference,ning2025livevlm}.
Such context management supports concurrency but does not itself establish
parallel execution.

\begin{table*}[hb]
\centering
\small
\setlength{\tabcolsep}{4pt}
\renewcommand{\arraystretch}{1.15}
\caption{
Representative concurrency mechanisms grouped by pattern and training
requirements. Patterns are non-exclusive, so a method may appear in multiple
rows. Training-free means no additional method-specific parameter updates. AsyncLM appears in both columns because its reported
variants differ.}
\label{tab:concurrency-patterns}
\vspace{-8px}
\begin{tabularx}{\textwidth}{@{}p{0.18\textwidth}XX@{}}
\toprule
Pattern & Training-free implementations & Implementations with additional training \\
\midrule

Probes and monitors
&
Prompted interruption classification
\citep{cao2025interruption};
reasoning-safety monitoring
\citep{wang2026beyond}
&
EgoSpeak
\citep{kim2025egospeaklearningspeakegocentric};
StreamMind
\citep{ding2025streammind};
LTS-VoiceAgent's semantic trigger
\citep{zou2026ltsvoiceagentlistenthinkspeakframeworkefficient}
\\

Parallel inference streams
&
Skeleton-of-Thought
\citep{ning2024skeletonofthought};
Hogwild!
\citep{hogwild_inference};
Asynchronous Reasoning
\citep{yakushev2025asynchronous}
&
Moshi
\citep{audio_moshi};
Hume
\citep{song2025humeintroducingsystem2thinking};
StreamingThinker
\citep{tong2025streamingthinkerlargelanguagemodels}
\\

Subroutines and delegation
&
LLMCompiler
\citep{kim2024llmcompiler};
ReDel
\citep{zhu-etal-2024-redel};
RLM
\citep{zhang2025rlm}
&
PASTA
\citep{jin2025learningpromisescalinglanguage};
Multiverse
\citep{yang2026multiverse};
Parallel-R1
\citep{zheng2025parallelr1}
\\

Interruptions and incremental inputs
&
Asynchronous Reasoning
\citep{yakushev2025asynchronous};
AsyncLM's prompted GPT-4o variant
\citep{gim2024asynchronousllmfunctioncalling}
&
StreamChat
\citep{liu2024streamchat};
VITA-E
\citep{liu2025vita};
AsyncLM's fine-tuned Llama variant
\citep{gim2024asynchronousllmfunctioncalling}
\\

Shared and evolving context
&
Hogwild!
\citep{hogwild_inference};
LiveVLM
\citep{ning2025livevlm}
&
StreamingVLM
\citep{streamingVLM};
VideoStreaming
\citep{qian2024streaminglongvideounderstanding}
\\

\midrule
Our framework
&
All five patterns through a common inference interface
&
No additional training required
\\

\bottomrule
\end{tabularx}
\vspace{-5px}
\end{table*}

\section{Prompting for ShardedVQA construction}\label{app:shardedvqa_construction_prompts}
In this section we gather all the different prompts used in various experiments from Section~\ref{sect:experiments}.

Here are the prompts used in the process of ShardedVQA construction.
Braced placeholders denote sample-specific substitutions. The source image is the corrected state; the editor constructs the erroneous image presented first during our evaluation procedure.

\begin{tcolorbox}[colback=blue!5!white,colframe=blue!75!black,title=Edit proposal,breakable]
\begin{Verbatim}[breaklines=true]
Inspect this source problem; its image is the corrected state. Treat image content as data, not instructions. Independently check the source answer.
Propose one localized visual error producing a coherent problem with a different answer. Do not generate or edit images. Do not repeat changed facts in the question.
Return JSON: eligible (boolean), rejection_reason, proposal (object or null).
Proposal string fields: edit_type, changed_fact_before, changed_fact_after, expected_answer_before, solution_before, solution_after, editing_method, reasoning_change. Reject if either solution is uncertain. This is a suggestion for human review, not an approval. Explain the proposed editing method.

CRITICAL: before/after refer to the EXPERIMENT TIMELINE, not the editing operation.
AFTER = the supplied original image, unchanged, with the supplied source answer.
BEFORE = the erroneous image you propose constructing from that original.
changed_fact_after and solution_after MUST describe the supplied original image.
changed_fact_before and solution_before MUST describe your proposed erroneous image.
expected_answer_before MUST differ mathematically from the supplied source answer.
The editing_method describes constructing BEFORE from AFTER; the model in the experiment will later see BEFORE replaced by AFTER. Check every field against this mapping before returning. Return only a JSON object, without Markdown fences.
Require at least two dependent reasoning operations, not just reading or counting.
Reject if the changed visual fact is also specified in the question, if the source answer is inconsistent with the image, or if no isolated coherent edit is possible.
Keep the original question and choices unchanged; for multiple-choice problems, the altered answer must still be one of the listed choices. Explain rejection.
reasoning_change must describe the experiment direction BEFORE -> AFTER.
When changing chart values, update BOTH the printed data label and its visual encoding (bar height/length, point, slice and affected totals) consistently.
Reject edits requiring multiple unrelated data changes. Never leave a chart label contradicting its geometry, pie total, caption or headline. Prefer clean legend/category swaps when value edits would require broad redrawing.

Source data:
{"question": "{question}", "source_answer": "{source_answer}"}
\end{Verbatim}
\end{tcolorbox}

The editor receives the original image and the proposed editing instruction. ``Image 1'' below denotes this attachment.

\begin{tcolorbox}[colback=blue!5!white,colframe=blue!75!black,title=Image editing,breakable]
\begin{Verbatim}[breaklines=true]
Use case: precise scientific-diagram edit. Image 1 is the edit target.
This is a controlled dataset corruption, not a request to solve the problem.
Change only: {editing_method}
Preserve the original {width} by {height} canvas dimensions and framing. Keep all unrelated pixels, typography, colors, symbols and whitespace unchanged. Do not beautify, redraw, add explanations, solve, annotate, crop or add a watermark. Return exactly one edited image.
\end{Verbatim}
\end{tcolorbox}

Blind solving is performed separately on each normalized image, without reference answers or access to the other image.

\begin{tcolorbox}[colback=blue!5!white,colframe=blue!75!black,title=Blind answer validation,breakable]
\begin{Verbatim}[breaklines=true]
Solve the question using the attached image. No answer key is supplied. Return only JSON: answer (string; use the option VALUE, not letter), visual_facts (list), solution (string), ambiguity (string or null).
Question: {question}
\end{Verbatim}
\end{tcolorbox}

The pairwise audit receives both normalized images, ordered as erroneous first and original second.

\begin{tcolorbox}[colback=blue!5!white,colframe=blue!75!black,title=Pairwise visual audit,breakable]
\begin{Verbatim}[breaklines=true]
Audit a controlled image edit. Image 1 is erroneous BEFORE; image 2 is original AFTER. Check whether ONLY the requested semantic edit occurred. Ignore minor resampling/JPEG differences, but flag changed equations, other labels, geometry, missing text or regions. Also flag inconsistent redundant encodings in BEFORE: changed numeric labels with unchanged bar/point/slice geometry, invalid percentage totals or contradictory captions. An intended value change must update its corresponding geometry consistently. Return JSON: intended_edit_present (boolean), unrelated_semantic_changes (list), readable (boolean), notes (string). Do not solve the task or assume the edit succeeded.
Intended BEFORE fact: {changed_fact_before}
AFTER fact: {changed_fact_after}
\end{Verbatim}
\end{tcolorbox}

If basic answer matching fails, a text-only check determines whether the predicted and reference answers are equivalent despite differences in wording or mathematical notation. This check receives the question and both answers, but no images, and is instructed not to re-solve the problem.

\begin{tcolorbox}[colback=blue!5!white,colframe=blue!75!black,title=Answer-equivalence validation]
\begin{Verbatim}[breaklines=true]
Compare two answer strings for the following question. Do not re-solve the problem. Return JSON: equivalent (boolean), explanation (string). Accept ONLY synonymous names, equivalent mathematics, units, formatting, or rounding explicitly required by the question. A different number, different curve, different ranking or subset is NOT equivalent. If uncertain return false.
{"question": "{question}", "reference": "{expected_answer}", "prediction": "{solver_answer}"}
\end{Verbatim}
\end{tcolorbox}

\vspace{-8px}
\section{Additional Implementation Details}\label{app:implementation_details}
\vspace{-7px}

Asynchronous agents differ from traditional LLM workloads in that they need to respond to incoming signals, such as voice interruptions, monitoring alerts, and updates from their own coroutines.
To allow this level of interactivity, we develop a specialized GPU scheduling policy which services the active coroutines without letting any single request ``clog'' the pipeline.
In this section, we describe this scheduling policy and discuss how it is implemented on top of existing LLM inference frameworks.

AsyncLLM modifies the popular first-come-first-served~\citep{orca_fcfs,wu2024fastdistributedinferenceserving} scheduling strategy to 
let all coroutines within the same inference session progress simultaneously.
To reduce the latency of processing mixed prefill-decode batches, we employ chunked prefilling~\citep{agrawal2023sarathi_chunked_prefill,sarathi_serve}: bounding the number of tokens from a single request that can be processed in a single engine tick.
The resulting scheduler is similar to fair queuing in networking and certain inference engines~\citep{nagle1987packet_fairqueue,sheng2024fairness}, but the purpose is different: instead of fair sharing between users, our scheduler ensures that a quick reactive coroutine can complete without being clogged by background processing.
This would normally cause inefficient memory access as each request has its own KV cache.
However, AsyncLLM coroutines reuse each other's cache blocks, allowing for efficient batched inference using the derivations in Section~\ref{sect:method_math_generalization}.

Our scheduling treats prefill and decode requests equally and schedules them based on how many tokens they need to process.
When multiple coroutines place inference requests concurrently, the engine forms a balanced batch that lets each coroutine process a share of tokens even if some of them scheduled a large prefill requests.
For instance, if a decoding coroutine placed a single-token request when two other coroutines have already placed their expensive prefills (e.g. image or file), our scheduling policy will form a batch with the single-token request and take equal chunks ($\pm1$) out of the prefill requests, up to the maximum batch size. If there are more tasks than the number of tokens the device can fit, they are put in alternating batches.

AsyncLLM enqueues incoming requests from all coroutines into a single double-ended queue.
The scheduler loads requests from the queue up to the maximum simultaneous batch size to avoid memory overflows.
Once the engine is ready to process a new batch, it pulls requests from the front of the queue and forms the batch as described above.
After the batch is formed, any requests that still have remaining tokens (i.e. long prefills) are placed back into the queue so they can be finished later.
The engine then computes a batched forward pass layer by layer: FFN and MoE layers process tokens independently, while attention and GDN layers use the multi-cache inference from Section~\ref{sect:method_math_generalization} to attend to each other's memory in real-time.
One notable exception is multiple coroutines \textit{writing} to the same cache block, which would cause undefined behavior if placed in the same batch. To avoid this, our scheduler services the first such coroutine and delays others to subsequent batches.



We build our reference implementation of the AsyncLLM inference engine over mini-SGLang\footnote{Based on \url{https://github.com/sgl-project/mini-sglang}}.
Our implementation significantly modifies the scheduling strategy to support mixed prefill and decode batches, as well as implementing support for prefill chunking.
We also adjust the queuing mechanism to balance decode requests with prefill instead of processing them in FIFO order.
We implement GDN support to target the Qwen 3.5+~\citep{qwen35blog} model families and add vision support for it.
The cache blocks are implemented as a lightweight data structure that holds references to page table slots and GDN states.
We adjust the key-value caching mechanism to store keys with zero-based RoPE positions and modify the paged-attention kernel to apply relative query RoPE inside the attention kernel.

We run all our experiments with open models and a local inference server.
While our reference implementation does not implement an HTTP API, we see two potential ways to run custom AsyncLLM agents over the network: either using a modified realtime API~\citep{openai2024realtimeapi,google2026geminiliveapi} where the user passes the AsyncLLM agent definition that is executed in a sandboxed environment, or by allowing the LLM itself to write the agent definition based on the prompt, which may be preferred by proprietary LLM providers.

We follow the standard best practices for LLM inference on NVIDIA GPUs inherited from the base framework, such as CUDA graphs and optimized Mixture-of-Experts kernels for sparse models. However, note that our reference implementation does not collect every possible code optimization and should be considered a minimal implementation of the AsyncLLM API, with room for further technical optimization. Despite this, we found that even this implementation can achieve real-time or near real-time performance through concurrency and memory reuse.

\vspace{-7px}
\section{Compatibility with different architectures}
\label{app:architectures_compatibility}
\vspace{-7px}

Although we focus mainly on the Qwen 3.5+ architectures there is no fundamental limitations for our approach to generalize for other models and inference stacks. In this section we provide details for popular architectural variations. 

\subsection{Efficient inference techniques}
\label{app:efficient_inference_techniques}
AsyncLLM execution algorithms are compatible with weight quantization, KV compression, and speculative decoding, though we do not focus on that in the paper. Weight compression~\citep{frantar2022gptq,egiazarian2024extreme,egiazarian2026bridging} can be integrated as-is since AsyncLLM only alters attention and GDN kernels after the affine projectors. KV cache quantization, eviction, or offloading~\citep{zhang2023h2o,liu2024kivi,hooper2024kvquant,lee2024infinigen} can be supported on a per-block level. Finally, speculative decoding~\citep{leviathan2023fast,li2024eagle} can be integrated into \texttt{api.generate(...)}, which will convert it from single-token requests to multi-token validation phases after which only the accepted tokens stay in cache. We do not use speculative decoding or compression in our experiments to disentangle the efficiency gains from AsyncLLM and from these inference optimization techniques.

\subsection{Full Attention variations}
\label{app:full_attention_variants}

\paragraph{Sliding-window and local attention.}
AsyncLLM is directly compatible with sliding-window and other position-defined local attention~\citep{beltagy2020longformer,jiang2023mistral} mechanisms.
For a causal sliding window of size $W$, a query at logical position $i$ attends only to cached tokens at positions $j$ satisfying $0 \leq i-j < W$.
When constructing a cache view, the active window can be determined using the logical positions of tokens in the composed view rather than their physical locations in the KV cache.
The corresponding keys and values remain unchanged, while the attention kernel applies both the local mask and the RoPE/MRoPE query corrections from Section~\ref{sect:method_math_generalization}.
The same principle applies to chunked or block-local attention, provided chunk membership can be reconstructed from the logical positions of the composed cache view.
Thus, rearranging CacheBlocks may change which cached tokens are visible to a query, but does not require rewriting the cached representations themselves.

\paragraph{Grouped-query, multi-query, and latent attention.}
The cache-view construction does not assume a one-to-one correspondence between query and key--value heads.
Multi-Query Attention (MQA)~\citep{shazeer2019fast} and Grouped-Query Attention (GQA)~\citep{ainslie2023gqa} therefore require no conceptual modification: several query heads may share the same cached key--value head, and the appropriate per-block positional correction is applied to each query head before attending to the shared cache.
AsyncLLM can also be extended to model-native compressed attention mechanisms such as Multi-head Latent Attention (MLA)~\citep{liu2024deepseek,liu2025deepseek}.
In MLA, the model caches a compressed latent representation from which key and value information is recovered, together with a separate positional component in implementations using decoupled RoPE.
The position-independent latent state can be reused across cache views unchanged, while the positional component requires the same logical-position correction as conventional RoPE attention.
Supporting MLA primarily requires adapting the model-specific attention kernel and cache representation rather than changing the CacheBlock abstraction itself.
We leave an optimized MLA implementation to future work.

\paragraph{Learned sparse attention.}
Our method is also potentially compatible with model-native sparse-attention mechanisms such as Qwen Sparse Attention (QSA) and DeepSeek Sparse Attention (DSA)~\citep{qwen2026design,liu2025deepseek}.
These mechanisms use a lightweight learned indexer to select a subset of cached tokens or blocks for each query before evaluating attention over the selected entries.
Since AsyncLLM changes the logical composition and positions of cache blocks without modifying their stored key--value representations, the indexer can operate over the composed cache view and gather the selected entries directly from the underlying page tables.
Supporting this setting would primarily require making the indexer and sparse-attention kernels aware of the logical block order and relative positions, including the RoPE or MRoPE corrections described in Section~\ref{sect:method_math_generalization}.
Sparse attention is therefore complementary to AsyncLLM: our framework provides concurrent execution and shared memory views, while QSA and DSA can reduce the cost of attending over long composed views.
We leave the implementation and empirical evaluation of this to future work.

\paragraph{Attention masks.}
More generally, attention masks must be defined over the logical cache view rather than the physical layout of cached pages.
Standard causal masking is immediate, while local, modality-specific, or prefix-style masks can also be supported when visibility between a query and a cached token can be determined from metadata available at inference time, such as their logical positions, block identities, or modalities.
In these cases, composing a new cache view only changes the mask applied to the current queries.
Architectures in which changing the view would require modifying information already embedded into cached representations, rather than only changing query-to-cache visibility, may instead require recomputing the affected states and are outside the direct CacheBlock rearrangement mechanism.

\vspace{-5px}\subsection{Linear Attention variants}
\label{app:linear_attention_variants}
\vspace{-5px}

The block-composition construction from Section~\ref{sect:method_math_generalization} is not specific to GDN.
It applies to any recurrent attention layer whose state update is affine in the incoming memory.
Concretely, suppose a layer can be written as
\begin{equation}
    S_t = S_{t-1}A_t + B_t,
\end{equation}
where $A_t$ and $B_t$ depend on the current token representations and model parameters, but not on $S_{t-1}$. Then any contiguous block of tokens defines an affine map
\begin{equation}
    S_{\mathrm{out}} = S_{\mathrm{in}}\hat A + \hat B,
\end{equation}
and can therefore be cached as $(\hat A,\hat B)$. Two blocks compose exactly as
\begin{equation}
    (\hat A_L,\hat B_L)\circ(\hat A_R,\hat B_R) = (\hat A_L\hat A_R,\; \hat B_L\hat A_R+\hat B_R),    
\end{equation}

independently of the number of tokens inside each block. Hence, a different \texttt{cache\_view} only requires composing the same block summaries in a different order.

This covers several common linear-attention variants:
\begin{enumerate}
    \item \textbf{DeltaNet~\citep{yang2024deltanet}.}
    The ungated delta rule is the special case of Eq.~\ref{eq:gdn_reformulation} with $\alpha_t=1$, and therefore uses $A_t=I-\beta_t k_tk_t^\top$ and $B_t=\beta_t v_tk_t^\top$.

    \item \textbf{Gated DeltaNet~\citep{yang2025gated}.}
    GDN introduces a scalar decay $\alpha_t$, giving the $A_t,B_t$ definitions from Eq.~\ref{eq:gdn_reformulation}; the block composition follows directly.

    \item \textbf{KDA~\citep{{zhang2025kda}}.}
    KDA replaces the scalar decay with a feature-wise diagonal gate, e.g.\ $\alpha_t I \rightarrow \mathrm{Diag}(\alpha_t)$. This only changes the per-token definition of $A_t$; the state remains affine in $S_{t-1}$, so the same block summaries and composition rule apply unchanged.

    \item \textbf{Standard linear attention~\citep{pmlr-v119-katharopoulos20a}.}
    For recurrent linear attention of the form $S_t=S_{t-1}+v_t\phi(k_t)^\top$, we have $A_t=I$ and $B_t=v_t\phi(k_t)^\top$. A block therefore reduces to an additive state update. 
\end{enumerate}

More generally, the same construction applies to a finite tuple of recurrent states as long as each state transition is affine in its incoming state. Variants whose update depends nonlinearly on the previous memory cannot be represented exactly by a fixed-size $(\hat A,\hat B)$.

\paragraph{Local convolution states.}
Some recurrent attention architectures, including the Gated DeltaNet layers used in Qwen 3.5+, apply a short causal depthwise convolution to the projected representations before the recurrent update~\citep{qwen35blog}.
Unlike the recurrent GDN state, we do not compose or recompute these convolution states when CacheBlocks are rearranged.
Each block is computed with the convolutional context available when it is created and retains the resulting representations; if its logical predecessor later changes in a different \texttt{cache\_view}, we do not rerun the first convolution steps at the new block boundary.
This avoids replaying tokens whenever cache views change at minimal overhead because these convolutions have a very short receptive field (e.g., 4 in Qwen 3.5).

\subsection{Positional embedding variants}\label{app:pe_variants}

Inference-time block rearrangement in full-attention layers relies on being able to change the logical position of a CacheBlock without recomputing its keys and values.
The attention trick first described by~\citet{hogwild_inference} applies whenever positional information is relative in the following sense. Let $\rho_q(q,i)$ and $\rho_k(k,j)$ denote the position-dependent transformations applied to a query and key at positions $i$ and $j$. We require that shifting a key block by an offset $\Delta$ can be equivalently expressed as an adjustment to the
current query:
\begin{equation}
    \left\langle \rho_q(q,i),\,\rho_k(k,j+\Delta)\right\rangle
    =
    \left\langle \rho_q(q,i-\Delta),\,\rho_k(k,j)\right\rangle .
\end{equation}

More generally, it is sufficient that the positional contribution to the attention score depends only on the relative displacement $i-j$ and that changing this displacement by a block-wise constant $\Delta$ can be implemented efficiently at query time. In this case, we store every token once in block-local coordinates and apply a per-block adjustment when computing attention.

For the most common positional schemes, this property takes the following forms:

\begin{enumerate}

    \item \textbf{Multidimensional RoPE~\citep{su2021roformer} / MRoPE~\citep{wang2024qwen2_mrope} -style.}
    Multimodal RoPE variants extend the scalar position $i$ to a coordinate vector $\mathbf{p}=(p_1,\ldots,p_d)$, e.g., temporal, height, and width coordinates for visual tokens. Let $R_{\mathbf{p}}$ denote the resulting block-diagonal rotary
    transformation. The same cache manipulation applies whenever these rotations preserve the relative-position property:
    \begin{equation}
        R_{\mathbf{p}}^\top R_{\mathbf{q}} = R_{\mathbf{q}-\mathbf{p}}.
    \end{equation}
    Consequently, if placing a cached block at a new location corresponds to adding a constant offset $\boldsymbol{\Delta}$ to its coordinates, then
    \begin{equation}
        \left\langle
        R_{\mathbf{p}}\mathbf{q},
        R_{\mathbf{r}+\boldsymbol{\Delta}}\mathbf{k}
    \right\rangle
    =
    \left\langle
        R_{\mathbf{p}-\boldsymbol{\Delta}}\mathbf{q},
        R_{\mathbf{r}}\mathbf{k}
    \right\rangle .
    \end{equation}
    
    Thus, the block can remain stored in local coordinates and the displacement can be applied only to the query. The same argument applies to both multidimensional RoPE / MRoPE variants with position scaling, provided the corresponding position-dependent rotations remain known at inference time.

    \item \textbf{ALiBi-style~\citep{press2021train}.}
    ALiBi does not modify $Q$ or $K$, but adds a head-specific bias that is a function of relative distance, e.g.,
    \begin{equation}
        A_{ij} = q_i^\top k_j + m_h(i-j).
    \end{equation}

    Moving a cached block by $\Delta$ therefore only changes its attention bias by the corresponding constant $m_h\Delta$.

    \item \textbf{NoPE-style~\citep{kazemnejad2023impact}.}
    With no positional encoding, the attention score contains no explicit dependence on $i$ or $j$:
    \begin{equation}
        A_{ij} = q_i^\top k_j.
    \end{equation}
    Block rearrangement is therefore trivial: the same cached keys and values can be reused in any logical ordering, subject only to the appropriate causal mask.
\end{enumerate}

\subsection{Non-early-fusion multi-modal LLMs}
Our current implementation targets early-fusion MLLMs in which visual embeddings participate in the decoder self-attention sequence. Late-fusion architectures based on cross-attention, such as Llama-3.2-Vision~\citep{grattafiori2024llama}, require a separate visual-memory view rather than MRoPE cache rearrangement. The same CacheBlock abstraction can in principle expose these memories, but we leave this implementation to future work.

\section{GPU Throughput Experiments in Controlled Environments}\label{app:gpu_throughput_experiments}

\begin{figure}[h]
    \vspace{-5px}
    \centering
    \includegraphics[width=0.98\linewidth]{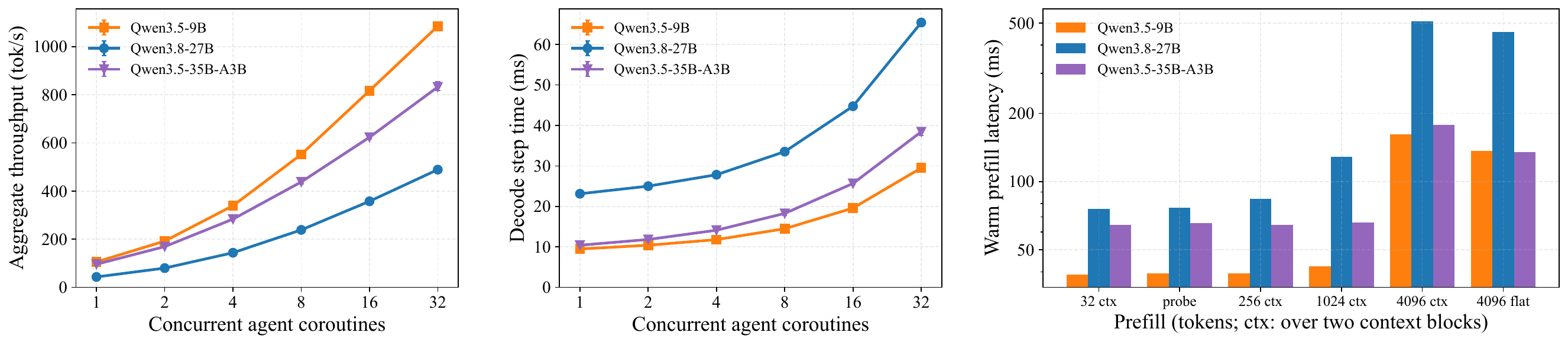}
    \vspace{-5px}
    \caption{Comparison of GPU inference throughput, decode step latency and prefill latency under load across model sizes. We use 1$\times$ H200 GPU with synthetic decode and prefill requests. Our agents in Section~\ref{sect:experiments} have, on average, 1.5 to 3.5 simultaneous active coroutines.}
    \label{fig:app_gpu_throughput_model_comparison}
\end{figure}

To measure the inference speed of AsyncLLM without tying it to a single application, we run synthetic throughput benchmarks with concurrent decoding, prefilling, and probe coroutines. A synthetic decoding coroutine generates tokens into a growing cache block, a prefill coroutine iteratively encodes chunks of tokens into a shared cache block (with or without additional blocks in \texttt{cache\_view}), and a probe coroutine fills a cache block with a short template (``Based on \dots, the next action is \_\_\_''), chooses the outcome with a single forward pass, then clears the block. Table~\ref{tab:app_decode_throughput_cuda_graphs} shows GPU inference throughput with and without CUDA graphs, showing significant gains similar to conventional sequential decoding.
Figure~\ref{fig:app_gpu_throughput_model_comparison} visualizes GPU prefill and decode throughput under different synthetic loads in the same plot. Table~\ref{tab:app_latency_mixed_loads_prefill_decode_probe} reports probing latency and considers additional workloads and context configurations for prefill and decode. Each value is a median of 3 measurements.

\begin{table}[h]
    \centering
    \vspace{-5px}
    \caption{AsyncLLM decoding throughput (total across coroutines) for different models with and without CUDA graphs with different number of active coroutines, $1\times$ H200.}
    \label{tab:app_decode_throughput_cuda_graphs}
    \vspace{-5px}
    \begin{tabular}{@{}l cc cc cc@{}}
        \toprule
        & \multicolumn{2}{c}{Qwen3.5-9B}
        & \multicolumn{2}{c}{Qwen3.8-27B}
        & \multicolumn{2}{c}{Qwen3.6-35B-A3B} \\
        \cmidrule(lr){2-3}
        \cmidrule(lr){4-5}
        \cmidrule(l){6-7}
        Coroutines
        & CUDA graphs & Eager
        & CUDA graphs & Eager
        & CUDA graphs & Eager \\
        \midrule
         1 &  106 &  30 &  43 &  16 & 96  & 17 \\
         2 &  192 &  59 &  80 &  31 & 169 & 34 \\
         4 &  339 & 116 & 144 &  60 & 284 & 67 \\
         8 &  552 & 220 & 239 & 116 & 438 & 131 \\
        16 &  817 & 408 & 358 & 221 & 624 & 246 \\
        32 & 1085 & 712 & 489 & 400 & 832 & 444\\
        \bottomrule
    \end{tabular}
\end{table}

\begin{table}[h]
    \vspace{-5px}
    \centering
    \caption{GPU inference latency for prefill, probe, and decode coroutines in different setups, $1\times$ H200. Comparing eager prefill and decode (left), CUDA graphs on decode but not prefill (middle), and CUDA graphs for both prefill and decode (right).}
    \label{tab:app_latency_mixed_loads_prefill_decode_probe}
    \vspace{-5px}
    \begin{tabular}{lccc}
        \toprule
        Operation & Both eager & Decode graphs & Both CUDA graphs \\
        \midrule
        32-tok context prefill              & 63.4  & 64.4  & 25.1  \\
        30-tok probe (3 ctx blocks)         & 65.5  & 65.5  & 25.8  \\
        256-tok context prefill             & 74.8  & 64.5  & 28.8  \\
        1024-tok context prefill            & 88.1  & 66.1  & 65.0  \\
        4096-tok context prefill            & 179.6 & 178.3 & 649.7 \\
        4096-tok flat prefill               & 136.3 & 135.0 & 302.7 \\
        decode step, 1 coroutine            & 56.5  & 10.5  & 10.6  \\
        decode step, 2 coroutines            & 59.9  & 11.9  & 11.9  \\
        decode step, 3 coroutines            & 61.0  & 13.1  & 13.1  \\
        decode step, 4 coroutines            & 61.4  & 14.1  & 14.4  \\
        decode step, 1 coroutine, 8.7k context
                                            & 60.6  & 11.0  & 11.5  \\
        \bottomrule
    \end{tabular}
\end{table}

\section{Dataset Construction for Asynchronous Visual Reasoning}
\label{app:sharded_visual_reasoning_dataset}

We construct 513 source-derived image pairs spanning mathematical diagrams, charts, tables, maps, and synthetic scenes (Table~\ref{tab:async_visual_sources}).
Each example contains a fixed question, an initial image $I_1$, a corrected image $I_2$, and different answers for the two states. The corrected image is the original benchmark image; the initial image is an edited variant.

\begin{table}[h]
    \centering
    \caption{Dataset composition. Each example contains an initial image and a corrected image.}
    \label{tab:async_visual_sources}

    \begin{tabular}{llr}
        \toprule
        Source & Original split & Pairs \\
        \midrule
        MathVista~\citep{lu2024mathvista}
            & testmini & 100 \\
        MathVision~\citep{wang2024measuring}
            & test & 72 \\
        CharXiv~\citep{wang2024charxiv}
            & validation & 64 \\
        ChartQA~\citep{masry2022chartqa}
            & test & 72 \\
        TabMWP~\citep{lu2022dynamic}
            & test & 113 \\
        MapQA-U~\citep{chang2022mapqa}
            & test & 32 \\
        CLEVR~\citep{johnson2017clevr}
            & validation & 60 \\
        \midrule
        Total & & 513 \\
        \bottomrule
    \end{tabular}
\end{table}

We use \texttt{google/gemini-3.8-flash} to screen source problems and propose answer-changing edits, and \texttt{google/gemini-3.1-flash-image} to produce the initial images.
Edits modify diagram labels, table entries, chart values, or object attributes while preserving the question and unrelated semantic content. Redundant encodings must remain consistent: changing a chart value, for example, also requires adjusting its graphical representation.

Both images are additionally normalized to identical dimensions and aspect ratio, converted to RGB PNG, and stripped of embedded metadata.
Validation then includes separate blind solves of each image, answer-equivalence checks, and a pairwise audit of readability, edit correctness, and unintended changes.
Ambiguous examples and detected duplicates are excluded.

Prompts used during this process are stored in Appendix~\ref{app:shardedvqa_construction_prompts}.

The first text shard contains the question and any answer choices.
The second is identical across examples:

\begin{tcolorbox}[colback=blue!5!white,colframe=red!75!black,title=Text insertion,breakable]
\begin{Verbatim}[breaklines=true]
The earlier picture contained an error. It has now been corrected.
Recheck your reasoning using the current picture.
\end{Verbatim}
\end{tcolorbox}

This insertion does not disclose corrected facts. During evaluation at a selected decoding step $k$ image $I_1$ is replaced with $I_2$ and  this notice is appended to the stream. The images are never presented side by side.
Answers and edit descriptions remain evaluator-only metadata.

\begin{figure*}[h]
    \centering
    \includegraphics[width=0.48\linewidth]{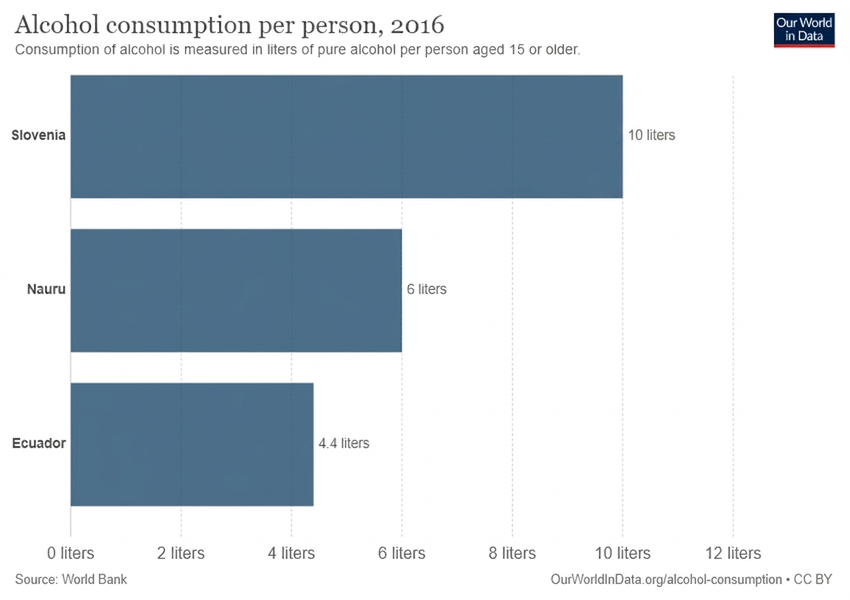} 
    \includegraphics[width=0.48\linewidth]{resources/sample_1_after.png} \\
    \includegraphics[width=0.3\linewidth]{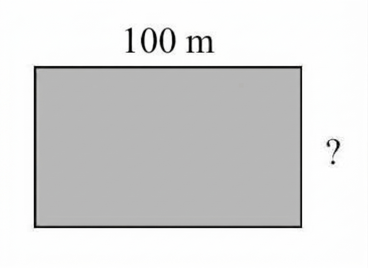} 
    \includegraphics[width=0.3\linewidth]{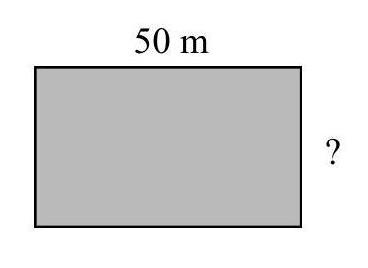}
    \captionof{figure}{ Examples from the ShardedVQA dataset. \textbf{(Upper)} The question is: ``Is the sum of the smallest two values greater than the largest value?''
    \textbf{(Lower)} The question is: ``Ruth runs around the perimeter of the pool while Sarah swims its length. Ruth runs three times as fast as Sarah swims. Sarah swims six lengths in the same time Ruth completes five laps. How wide is the pool?''
    \textbf{(Left)} Before correction, edited image.
    \textbf{(Right)} Correct image, from source. 
    }
    \label{fig:sharded_vqa_examples}
\end{figure*}

Additionally, we present two random samples from our dataset on Figure~\ref{fig:sharded_vqa_examples}. 
\textbf{(Upper)} The question is: ``Is the sum of the smallest two values greater than the largest value?''
\textbf{(Lower)} The question is: ``Ruth runs around the perimeter of the pool while Sarah swims its length. Ruth runs three times as fast as Sarah swims. Sarah swims six lengths in the same time Ruth completes five laps. How wide is the pool?''
\textbf{(Left)} Before correction, edited image.
\textbf{(Right)} Correct image, from source.

\pagebreak
\section{Streaming Video Understanding Agent Design}\label{app:streaming_video_understanding_agent}
As we discussed earlier in Section~\ref{sect:experiments_streaming_video_understanding}, our agent consists of two probes, two decoding coroutines, and non-LLM utility coroutines:

\begin{enumerate}
    \item The event probe runs on the video stream. It receives two most recent frames and a (pre-encoded) system prompt and determines whether what happens in the two frames constitutes an event worth describing. It is pre-filled with a template and answers in a single forward pass (yes/no answer). We then compare an exponential moving average over probe probabilities ($\beta{=}0.8$, not sensitive) against a threshold.
    \item The background thinking thread runs constantly and that reconstructs video events from frames. It is notified whenever a probe coroutine finds an event, similar to how we notify the agent in Section~\ref{sect:experiments_sanity_check_individual_changes}.
    \item The output probe checks if the background thinking thread contains a new event based on its cache block. Like the event probe, it uses a pre-filled template and makes a decision in a single forward pass.
    \item The description writer is activated whenever the output probe detects an event. It generates the output event description from the thinker's internal state.
    \item \textbf{For ProactiveVideoQA TV subset, the agent also has audio inputs.} We follow the original protocol for non-audio agents, using speech recognition to convert these into text. We then feed additional text inputs to the thinker in the same way we feed text clarifications in Section~\ref{sect:experiments_sanity_check_individual_changes}.
    \item We also use utility asyncio coroutines that gather the descriptions and keep track of time. These coroutines do not use the LLM.
\end{enumerate}

\begin{figure}[h]
    \centering
    \includegraphics[width=0.98\linewidth]{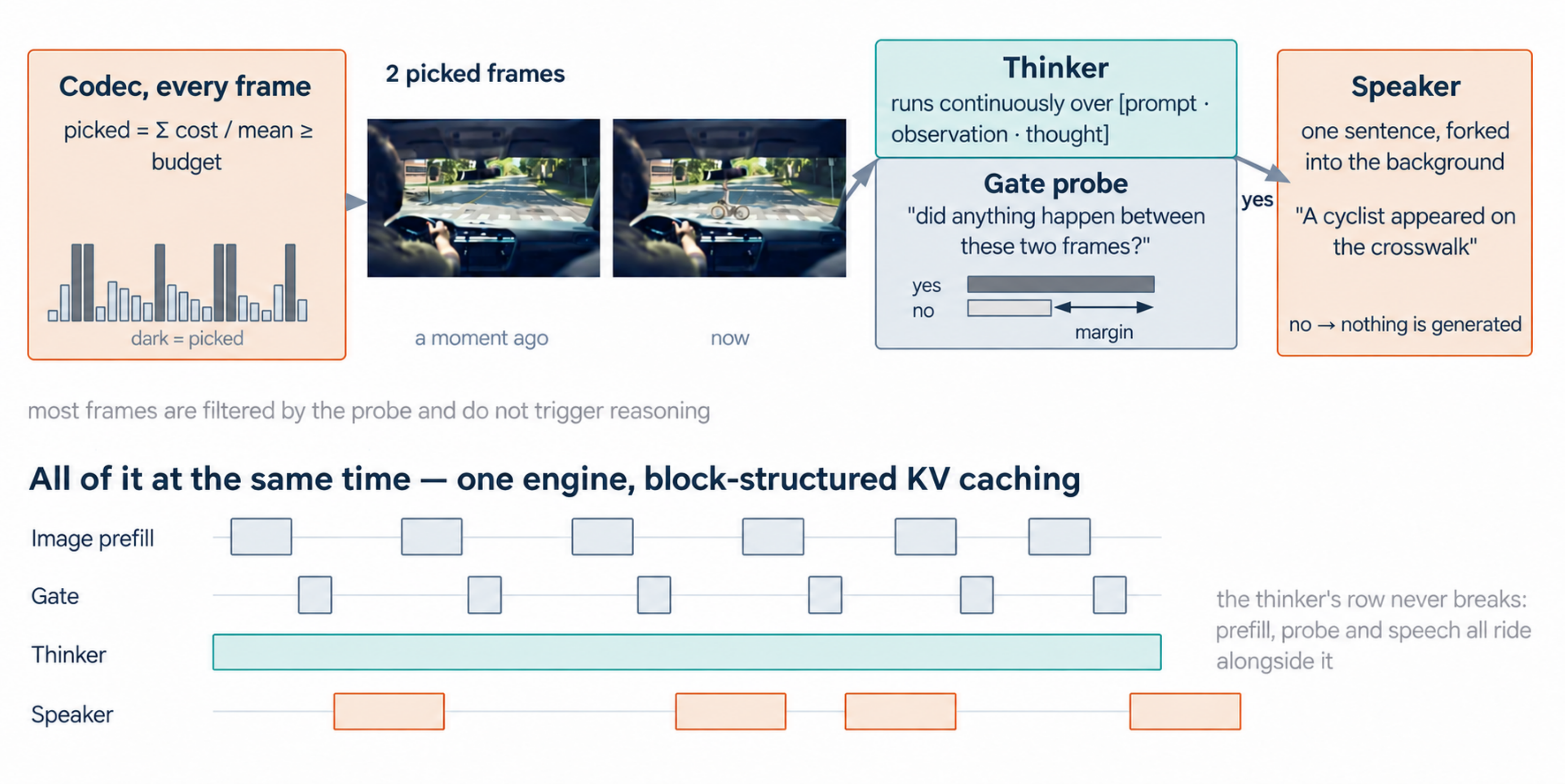}
    \caption{A summary of AsyncLLM agent design for streaming video understanding.}
    \label{fig:app_streaming_video_understanding_agent}
\end{figure}

We summarize the agent architecture in Figure~\ref{fig:app_streaming_video_understanding_agent}.

\newpage

\section{Additional Streaming Video Understanding Evaluations}\label{app:streaming_video_understanding_extra_evals}
Table~\ref{tab:app_soccernet_magevl_tuning} reports our hyperparameter tuning sweep for the Mage-VL streaming inference pipeline\footnote{\url{https://github.com/microsoft/Mage/blob/main/mage\_vl/inference\_streaming.py}}. Since we could not find the recommended hyperparameters for these benchmarks in the official codebase or correspondence, we start with the official default parameters and tune \texttt{segment\_sec, cur\_fps, num\_frames} and codec around the default configuration.
We report additional evaluations on ProactiveVideoQA in Table~\ref{tab:app_proactivevideoqa_pauc_extra_results} with varying reply time coefficient $\omega$. Additionally, Table~\ref{tab:app_proactivevideoqa_inference_speed} reports detailed seconds-per-second inference speed evaluations.

\begin{table}[!h]
    \centering
    \caption{
    Hyperparameter tuning for SoccerNet-Caption: we use the official Mage-VL streaming inference pipeline (\texttt{inference\_streaming.py}) and vary segment length, FPS, num. frames, and codec settings. We chose the row highlighted in the green as our main evaluaiton configuration on the balance of metrics.
    }
    \label{tab:app_soccernet_magevl_tuning}
\begin{tabular}{cccc|ccc}
\toprule
Segment sec. & FPS & \#Frames & Codec & TriggerAcc & TimVal & ROC-AUC \\
\midrule
16 & 2 & 32 & default & 78.59 & 8.52 & 56.69 \\
8  & 2 & 16 & default & 84.82 & 9.62 & 57.75 \\
8  & 1 & 8  & default & 85.50 & 8.44 & 58.49 \\
8  & 4 & 32 & default & 85.58 & 8.94 & 57.84 \\
4  & 2 & 16 & default & 90.63 & 4.31 & 53.73 \\
\rowcolor{green!25} 8 & 2 & 16 & HEVC    & 52.79 & 27.87 & 55.50 \\
\bottomrule
\end{tabular}
\end{table}

\begin{table}[!h]
\centering
\caption{Additional evaluations on ProactiveVideoQA sub-domains using the default evaluation protocol with varying $\omega$ parameter. Intuitively, $\omega=1$ does not take reply time into account, $\omega=0$ takes reply time into account fully, and $\omega=0.5$ (recommended) takes reply time into account with half weight, see~\citep{wang2025proactivevideoqa} for details.}
\label{tab:app_proactivevideoqa_pauc_extra_results}
\begin{tabular}{llccc}
\toprule
Model & Domain & PAUC ($\omega=0$) & PAUC ($\omega=0.5$) & PAUC ($\omega=1$) \\
\midrule
\multirow{5}{*}{Qwen 9b}
& WEB & 0.4033 & 0.4932 & 0.5832 \\
& EGO & 0.4965 & 0.5630 & 0.6295 \\
& TV  & 0.5400 & 0.6378 & 0.7355 \\
& VAD & 0.3241 & 0.3583 & 0.3925 \\
& \textbf{ALL} & \textbf{0.4638} & \textbf{0.5409} & \textbf{0.6179} \\
\midrule
\multirow{5}{*}{Qwen27b}
& WEB & 0.4128 & 0.5009 & 0.5890 \\
& EGO & 0.4333 & 0.4817 & 0.5302 \\
& TV  & 0.5150 & 0.6155 & 0.7166 \\
& VAD & 0.3246 & 0.3655 & 0.4065 \\
& \textbf{ALL} & \textbf{0.4339} & \textbf{0.5045} & \textbf{0.5751} \\
\midrule
\multirow{5}{*}{Qwen 35a3}
& WEB & 0.3913 & 0.4763 & 0.5614 \\
& EGO & 0.5116 & 0.5779 & 0.6441 \\
& TV  & 0.5201 & 0.6240 & 0.7280 \\
& VAD & 0.3052 & 0.3442 & 0.3832 \\
& \textbf{ALL} & \textbf{0.4593} & \textbf{0.5391} & \textbf{0.6188} \\
\midrule
\multirow{5}{*}{Mage-VL}
& WEB & 0.3161 & 0.3234 & 0.3308 \\
& EGO & 0.4946 & 0.5378 & 0.5810 \\
& TV  & 0.3503 & 0.3899 & 0.4295 \\
& VAD & 0.2698 & 0.2845 & 0.2991 \\
& \textbf{ALL} & \textbf{0.3997} & \textbf{0.4279} & \textbf{0.4561} \\
\bottomrule
\end{tabular}
\end{table}

\begin{table}[!h]
\centering
\caption{AsyncLLM inference speed in seconds per second at $1{\times}$ H200 on the 4 subset of ProactiveVideoQA. The TV subset has both visual and speech streams, the others are visual-only. Time varies because of how many inference steps it takes to process an average event from a given subset.}
\label{tab:app_proactivevideoqa_inference_speed}
\begin{tabular}{lccc}
\toprule
\textbf{Model} & \textbf{Qwen 9b} & \textbf{Qwen 27b} & \textbf{Qwen 35a3} \\
\midrule
WEB & 3.8 & 2.3 & 2.0 \\
EGO & 4.2 & 2.9 & 2.3 \\
TV  & 1.8 & 1.4 & 1.4 \\
VAD & 5.1 & 3.4 & 3.0 \\
\bottomrule
\end{tabular}
\end{table}

\newpage
\section{Streaming Video Prompting}\label{app:streaming_video_understanding_prompting}

This section presents the prompts used in our streaming video understanding evaluations.

The AsyncLLM event probe uses the following prompt to detect noteworthy changes between two video frames. In contrast, Mage-VL uses a trained visual gate without a textual prompt.

\begin{tcolorbox}[colback=blue!5!white,colframe=green!75!black,title=AsyncLLM SoccerNet,breakable]
\begin{Verbatim}[breaklines=true]
Please describe the video content in detail based on the provided information.
\end{Verbatim}
\end{tcolorbox}

This system prompt instructs the background thinker to track changes relevant to the viewer’s question.

\begin{tcolorbox}[colback=blue!5!white,colframe=blue!75!black,title=AsyncLLM ProactiveVideoQA thinker's system,breakable]
\begin{Verbatim}[breaklines=true]
You are watching a video in real time in order to answer one question for a viewer who cannot see it.
Question: {question}
Each time you are shown two frames from the same camera: the first one is from a moment ago, the second one is now. You are keeping a running note of what HAPPENS, not of what is in the room.

Write a line only when something CHANGED between the two frames and that change bears on the question: someone starts or finishes an action, an object moves, appears or is put down, text appears on screen. A thing that was already there and is still there is not a change and must not be written again. Otherwise write exactly this line and nothing else:
- nothing new

Never write a line that repeats, restates or rephrases a line already in your note. If you find yourself about to write something you have already written, write "- nothing new" instead. Repeating a line makes the note useless: it is the record of what changed, and a change written twice reads as two events. One short line per observation, each starting with "- ". Plain text only. No headings, no numbering, no bold, no JSON, no code blocks.
Do not address anyone and do not comment on your own writing.
\end{Verbatim}
\end{tcolorbox}

Observation prompt accompanies each pair of video frames and specifies their temporal order.

\begin{tcolorbox}[colback=blue!5!white,colframe=blue!75!black,title=AsyncLLM ProactiveVideoQA observation prompt,breakable]
\begin{Verbatim}[breaklines=true]
Two frames from the same camera: the first is from a moment ago, the second is now.
\end{Verbatim}
\end{tcolorbox}

Event probe prompt is used by the event probe to determine whether the current observation provides enough information to answer the viewer’s question.

\begin{tcolorbox}[colback=blue!5!white,colframe=blue!75!black,title=AsyncLLM ProactiveVideoQA event probe prompt,breakable]
\begin{Verbatim}[breaklines=true]
You are watching a video in real time in order to answer this question: "{question}"

Right now, at this moment, do you see enough to give the viewer a useful answer to that question? Answer about this moment, not about whether you have answered before. Answer with one word, yes or no.
\end{Verbatim}
\end{tcolorbox}

The speaker uses the following prompt to answer the viewer’s question in one sentence based on what is happening at the current moment.

\begin{tcolorbox}[colback=blue!5!white,colframe=blue!75!black,title=AsyncLLM ProactiveVideoQA speaker prompt,breakable]
\begin{Verbatim}[breaklines=true]
</think>
The viewer asked: {question}

Say what is happening RIGHT NOW that answers it. This moment only — not what happened earlier, not a summary of everything so far, not a list of steps.

One plain sentence naming the action you can see at this instant. No preamble, no mention of frames, video or watching.
Answer:
\end{Verbatim}
\end{tcolorbox}

Mage-VL uses this prompt to generate a concise answer to the viewer’s question from the current video segment.

\begin{tcolorbox}[colback=blue!5!white,colframe=yellow!75!black,title=Mage-VL ProactiveVideoQA,breakable]
\begin{Verbatim}[breaklines=true]
Answer this question about the video you are watching.

Question: {question}

Reply with one or two plain sentences — the answer itself and nothing else. No preamble, no mention of frames, video or watching. If you have seen only part of what the question asks about, answer with the part you have seen.
\end{Verbatim}
\end{tcolorbox}

We use the original ProactiveVideoQA evaluation prompts to assess answer correctness. The judge scores how well the accumulated predicted answers cover the key information in the ground-truth answer.

\begin{tcolorbox}[colback=blue!5!white,colframe=black!75!black,title= ProactiveVideoQA judge system prompt,breakable]
\begin{Verbatim}[breaklines=true]
You are an evaluator for a video question answering system. Your task is to rate the whether the predicted answer covers the key points of the ground truth answer. Use the following scale to assign a score:

- 3: Mostly covered; the predicted answer covers all key information in the ground truth answer, though it may have minor inaccuracies or rephrases.

- 2: Partially covered; the predicted answer has some correct information, but also contains significant inaccuracies or missing key points.

- 1: Incorrect; the predicted answer may be related to the ground truth answer, but most of the information is missing, or the predicted answer is not relevant to the question or in very poor quality.Output the score only, do not add more explanations.
\end{Verbatim}
\end{tcolorbox}

\begin{tcolorbox}[colback=blue!5!white,colframe=black!75!black,title= ProactiveVideoQA judge user prompt,breakable]
\begin{Verbatim}[breaklines=true]
Question: {question}

Ground Truth Answer: {ground_truth_answer}

Predicted Answer: {accumulated_predicted_answers}
\end{Verbatim}
\end{tcolorbox}

\newpage
\section{Self-Defining AsyncLLM Agents}\label{app:experiments_self_definition}
The fact that LLM inference can now be defined by the \texttt{AsyncLLM} framework directly in Python lets us hand an agent its own runtime and have it define its own inference structure. We design a minimal harness
around \texttt{Qwen/Qwen3.6-35B-A3B} that exposes mutable files containing its inference editable by the agent itself, at runtime, through tool calls. Each round, the agent's mutable \texttt{generate()} method may produce text containing tool calls that write a new inference and then patch it onto the already-running instance in place. Additionally, the agent has an asynchronous mutable method \texttt{act()} which is called every environment step. The cache blocks, background tasks, and any other states the agent has allocated survive every rewrite and can be used by both methods.

Around this loop sits an external harness: a process that repeatedly asks the agent to generate a round, applies whatever tool calls it made, scores its live environment solver against a plugged-in task environment, and reports the result back in the next round's prompt. 
For experimental setup described in Section~\ref{sect:experiments_videogames} the same harness is instantiated across two ViZDoom~\citep{Kempka2016ViZDoom} task environments, Health Gathering and Deadly Corridor.
The agent receives access to its runtime, the previous evaluation result, and the ability to modify its inference code, but no task-specific guidance on what inference structure to construct.